\documentclass{article}
\usepackage{preprint,times}
\usepackage{amsmath,amsfonts,bm}

\def\eqref#1{equation~\ref{#1}}
\def\Eqref#1{Equation~\ref{#1}}

\def\1{\bm{1}}

\DeclareMathAlphabet{\mathsfit}{\encodingdefault}{\sfdefault}{m}{sl}
\SetMathAlphabet{\mathsfit}{bold}{\encodingdefault}{\sfdefault}{bx}{n}

\usepackage{hyperref}
\usepackage{url}
\usepackage{booktabs}
\usepackage{amsmath,amssymb,amsthm}
\usepackage{multirow}
\usepackage{float}%
\usepackage{graphicx}
\usepackage{xcolor}
\usepackage{colortbl}
\definecolor{oursbg}{HTML}{E6EEF8}
\usepackage{enumitem}
\newcommand{\mk}[1]{MK#1}
\newcommand{\infllm}{InfLLM-V2}
\newcommand{\mcpm}{MiniCPM4.1-8B}
\newcommand{\seer}{SeerAttention}
\newcommand{\nosa}{NOSA}


\newcommand{\fitwidth}[1]{\resizebox{\ifdim\width>\linewidth\linewidth\else\width\fi}{!}{#1}}
\newtheorem{proposition}{Proposition}

\title{Lost in a Few Heads:\\ Diagnosing and Repairing Block Selection\\ in Sparse Attention}

\author{Wenhe Zhang\thanks{Correspondence to \texttt{wenhe.zhang@yale.edu}.}\\
Yale University
\And
Boxing Chen\\
Huawei Canada}

\begin{document}
\maketitle
\begin{abstract}
Sparse attention models read a few blocks of keys for every query at a fixed budget and can lose
much of their long-context retrieval accuracy at that budget. We ask where this loss is made. Using exact attention mass
selection at the shipped budget as an oracle, we show that the selector under-ranks the needle rather than
running out of budget. We find attention heads that miss the needle by running a calibration probe that teacher-forces the answer and reads two quantities in every head. The
first is the needle attention that the selector forfeits, and the second is the share of the answer's log-probability
that fixing the head's selection restores, integrated along the path from the shipped selector to full attention. Four
checkpoints of the \infllm{} family and \nosa{}-8B lose multi-key retrieval in the same three blind heads, which attend to
the needle while a mean-pooled key summary dilutes its single decisive token. Full attention in those heads or a 4-bit
scorer that keeps the shipped budget recovers most of the loss without training. On \mcpm{} at 64K, either fix lifts
multi-key retrieval accuracy from 54.1 to 97.5\%, level with full attention in every head. In production serving engines, full attention in those heads keeps 84 to 89\% of the time to first token (TTFT) that sparse
attention saves at 32K to 128K context length. Beyond the \infllm{} selector lineage, one unified rule flags attention heads above a threshold swept on each
checkpoint until they close most of the gap to full attention on calibration prompts. A similar small set of blind attention heads carry the loss for the SeerAttention-R gates, NSA, and DashAttention, whereas many such heads exist in MoBA.
On Qwen3-8B with the SeerAttention-R gates, full attention in 18 of 288 key-value heads lifts multi-key retrieval accuracy
from 29.8 to 97.7\%.
\end{abstract}

\section{Introduction}
\label{sec:intro}

\begin{figure}[t]
\centering
\includegraphics[width=\linewidth]{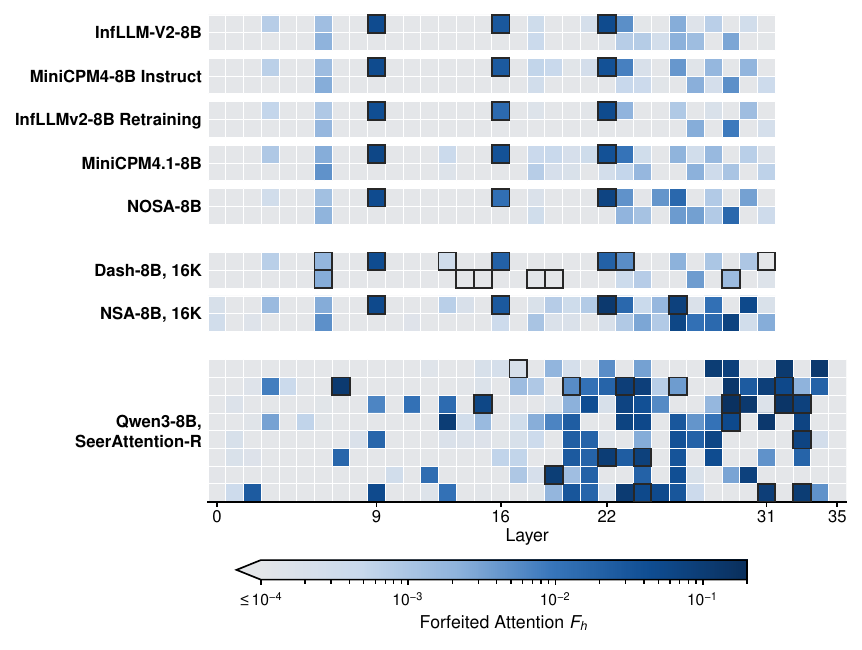}
\caption{\textbf{Where the selector forfeits the needle.} Per key-value head, the needle recall that the shipped
selector loses against exact mass, times the attention the head puts on the needle block, on calibration prompts at
32K, or 16K for Dash-8B and NSA-8B. The first group holds the five checkpoints of the MiniCPM4-8B lineage, with the
\infllm{} selector or the \nosa{} variant of it, head 0 above head 1. Dash-8B and NSA-8B share the backbone under other
selectors, and the last band is Qwen3-8B under the SeerAttention-R gate, heads 0 to 7 from the top. Outlined heads are
the heads that each model's fixes use, the three blind heads in the lineage and the heads of the threshold rule on the
other selectors. Scores at or below $10^{-4}$ share the lightest color.}
\label{fig:heads}
\end{figure}

Large Language Models (LLMs) are increasingly applied to long context tasks, including multi-turn chat conversations,
long document understanding, agentic workflows and multi-step reasoning
\citep{claudeopus55,deepseekv4,gemini38,kimik3,gpt6astra,glm53}. Full attention has a cost quadratic in the context
length. As context windows grow from 32K to 128K tokens and beyond, attention comes to dominate both the latency of the
prefill and the memory that the key-value cache holds \citep{flashattention,flashattention2,flashattention3,effattnsurvey}.

Sparse attention has become a promising alternative, now trained into released models \citep{effattnsurvey}.
Token-level designs score every past key with a lightweight indexer and keep the highest-scoring keys
\citep{dsa, glm5}. Block-level designs pool contiguous blocks of keys and select the top-$k$ whole blocks, which is
far cheaper to score and maps better onto the memory system \citep{nsa,moba}. In both, the attention that follows
selection is exact, and the selector runs at a budget fixed by the model producers.

A fixed budget hides a subtle failure. Because attention after selection is exact, a sparse model's accuracy loss
comes from the selection, and a budget that is adequate on average can still be spent on the wrong blocks for a
particular query. Sparsification can change what a model relies on in ways that aggregate accuracy hides
\citep{cfaudit}. On RULER's multi-key retrieval at 32K, \infllm{}-8B's own report puts its sparse mode far below the
same model run with full attention \citep{ruler, infllmv2}. We measure similar losses for \mcpm{} \citep{minicpm4}, for
the SeerAttention-R gates \citep{seerattentionr} and for MoBA \citep{moba}, all within their native context ranges.

Fixes that do not locate the loss are costly. Raising the budget in every head reads more blocks at every query and
gives back the savings that sparse attention was adopted for. Retraining the model or its selector is expensive, and a
selector trained to match another one learns its failures as well \citep{sparda}. We therefore ask where the loss is
made and answer in four steps. First, exact attention mass selection at the shipped budget, applied by token and by
head, measures how much accuracy the choice of blocks costs. It shows that the selector ranks the needle too low, not
that the budget is too small. Second, a calibration probe localizes the loss to heads. It teacher-forces the answer
and reads, in every head, the needle attention that the selector forfeits and the share of the answer that fixing that
head's selection restores. Then, a two-block model of pooled scoring organizes the misses that the probe measures.
Finally, training-free fixes act in the heads the probe finds, and we report what each fix gains and what it costs.

How few heads carry the loss is itself a measurement, and it depends on the selector. In the MiniCPM4-8B lineage, four
checkpoints of the \infllm{} family and \nosa{}-8B lose multi-key retrieval in the same three blind heads (Figure~\ref{fig:heads}). Beyond the
lineage, one general rule sets the heads on every checkpoint: a threshold on the probe's score swept until the flagged heads
close most of the gap to full attention on calibration prompts. Under the SeerAttention-R gates and NSA, a few heads among
several hundred or dozens carry the loss. Under DashAttention, a larger set led by the same three blind heads carry the loss, and under
MoBA, many such heads exist. We build the fixes
on \infllm{}-8B as an add-on that changes only the top-$k$ path of the flagged heads, extend them to the checkpoints
that share its selector and to \nosa{}-8B \citep{nosa}, and then to other selector designs
\citep{seerattention,seerattentionr,dashattention,nsa,moba}.

In summary, our contributions are as follows.
\begin{itemize}[leftmargin=1.2em,itemsep=1pt,topsep=2pt]
\item \textbf{A diagnostic for selection failures.} Exact mass selection at the shipped budget separates a ranking
failure from a budget shortfall. Applied by token and by head, it shows where the loss is made
(Section~\ref{sec:diagnosis}).
\item \textbf{A calibration probe with two readouts.} Teacher-forced passes give each head the needle attention its
selector forfeits, which explains the failure. They also give each head its fix gain, the integrated effect of fixing its selection on the
answer, which picks the heads to fix. On the MiniCPM4-8B lineage, both read-outs return the same three blind heads
(Section~\ref{sec:diagnosis}).
\item \textbf{A resolution account.} A two-block model shows how a score that grows with a window's mean logit misranks
a needle that exact mass keeps. It organizes the measured misses of the blind heads, and keys quantized to a few bits
keep the needle's rank (Section~\ref{sec:diagnosis}, Appendix~\ref{sec:theory}).
\item \textbf{Training-free fixes and their costs.} Full attention in the flagged heads, a 4-bit scorer at the shipped
budget and a larger budget in the flagged layers each recover most of the multi-key loss. Across 17 task types, the scorer at the decode
tokens is the safe default for an unknown workload. Against the dense model in the producers' engine, full attention
in the flagged heads keeps most of the time to the first token (TTFT) that sparse attention saves (Sections~\ref{sec:method},
\ref{sec:results} and~\ref{sec:selectors}).
\item \textbf{One rule across selector designs.} A threshold on the fix gain, swept per checkpoint to a fixed share of
the gap, flags a few heads under the SeerAttention-R gates and NSA, a larger set under DashAttention and many heads under
MoBA. Where few heads carry the loss, the flagged heads recover more than RazorAttention's echo heads and random heads of
the same count. Under the SeerAttention-R gate, the heads flagged on synthetic prompts also recover part of the loss on a
real long-context benchmark, and token-level selectors show no multi-key loss (Section~\ref{sec:selectors}).
\end{itemize}

\section{Related Work}
\label{sec:related}

\textbf{Trainable sparse attention.} Some recent works train the selector with the model. Selection is either over
blocks in NSA \citep{nsa}, MoBA \citep{moba}, \infllm{} \citep{infllmv2,minicpm4}, DashAttention \citep{dashattention},
\nosa{} \citep{nosa} and MiniMax Sparse Attention \citep{msa}, or over single tokens in DSA \citep{dsa} and LongCat
\citep{longcatlsa}. Others train only a selector for a frozen model, such as a block gate in \seer{}
\citep{seerattention,seerattentionr}, a distilled block router in SpotAttention \citep{spotattention} and a forecast
of the next layer's blocks in SparDA \citep{sparda}.

\textbf{What a block summary loses.} A block summary can hide the block's heaviest token \citep{hisa,locks,fathom},
which RBS-Attention \citep{rbs} calls mean dilution and rescues at prefill. \citet{optmoba} show that a centroid
router's signal-to-noise ratio falls with the block size. COBS \citep{cobs} shows that selecting the blocks of largest
attention mass lets block sparse attention match full attention and that mean-key selectors approximate it to first
order. \citet{oracleprefill} use the same oracle, averaged over heads, to separate the budget from the indexer's error.
Quest \citep{quest} bounds a block by the per-channel extremes of its keys. Multipole \citep{multipole}, Prism
\citep{prism} and COBS improve the block summary. LOCKS \citep{locks} ranks pages by a log-sum-exp mass from one
compact code per page and proves that a code shared by a page's keys cannot separate some pages that differ inside.
Proposition~\ref{prop:pool} gives both sides, the margin at which exact mass keeps the needle block and the margin at
which a mean score loses it to competitor blocks.

\textbf{Low-bit scoring.} Scanning a compressed copy of the keys is explored for dense models
\citep{doublesparsity,loki,sale,ffd,fathom,fier}. Twilight \citep{twilight} and SpeCache \citep{specache} rank a base
selector's picks or an offloaded cache with a low-bit copy of the keys. HISA \citep{hisa} and self-indexing attention
\citep{selfindexing} replace a deployed token indexer in every head.

\textbf{Finding the heads that matter.} A few heads carry retrieval in dense models \citep{retrievalhead,duoattention}.
DuoAttention learns which heads keep the full cache. RazorAttention \citep{razorattention} keeps fixed fractions of
heads ranked by echo and induction scores, which it measures on random tokens repeated several times. RTPurbo
\citep{rtpurbo} gives full attention at prefill and a trained indexer at decode to heads found with a needle probe.
AB-Sparse \citep{absparse} calibrates a block size per head, and \citet{dynamicheads} argues that the retrieving heads
vary within a generation. These criteria score a behavior of the head in a dense model. Our probe scores a deployed
selector's failure in each head against exact mass at the shipped budget and ranks heads by the effect of fixing that
failure on the answer, integrated along the path from the shipped selector to full attention
\citep{sundararajan2017ig,hanna2024faithfulness}, which at its start is an attribution patching estimate
\citep{nanda2023atp,syed2023atp,kramar2024atpstar} of the fix itself.

\section{Preliminary}
\label{sec:prelim}

\textbf{Full attention with grouped queries.} For one layer, query token $t$, query head $g$ and head dimension $d$,
softmax attention computes $\bm{o}^{(g)}_t = \sum_{j\leq t} a^{(g)}_{t,j}\, \bm{v}_j$ with
$a^{(g)}_{t,j} = \operatorname{softmax}_j(\bm{q}_t^{(g)\top} \bm{k}_j/\sqrt d)$, at a cost quadratic in the sequence
length $N$. Grouped Query Attention \citep{gqa} lets $G$ query heads share one key-value head.

\textbf{Block sparse attention.} Sparse attention chunks the keys into contiguous blocks
$\mathcal{B}_1,\mathcal{B}_2,\dots$, each of $L$ tokens. A block sparse selector gives each query a forced set
$\mathcal{F}$ of blocks it always reads, the initial block $\mathcal{B}_{init}$, a local window $\mathcal{B}_{local}$ and
the query's own block $\mathcal{B}_{query}$. It fills $K$ free slots with the blocks of highest score $c_b$,
$\mathcal{S}=\operatorname{TopK}_{b\notin \mathcal{F}}(c_b,K)$. Attention is then exact over the keys of
$\mathcal{F}\cup \mathcal{S}$, and the budget $|\mathcal{F}|+K$ is fixed. Block selectors differ in how the score $c_b$
is computed.

\textbf{Pooled-key selection.} \infllm{} \citep{infllmv2} scores blocks from mean-pooled keys over windows $W_m$ of $w$
consecutive keys. Each query head computes a softmax over the pooled keys, the heads of the group are summed and a
max-pool maps the window scores back onto blocks,
\begin{equation}
\bm{\bar{k}}_m=\frac{1}{w}\sum_{j\in W_m}\bm{k}_j,\qquad
c_b^{\mathrm{pool}}=\max_{m\in\mathcal W(b)}\;\sum_{g=1}^{G}
\operatorname{softmax}_m\!\left(\frac{\bm{q}_t^{(g)\top}\bar{\bm{k}}_m}{\sqrt d}\right),
\label{eq:pool}
\end{equation}
where $\mathcal W(b)$ is the set of windows that the kernel's max-pool assigns to block $b$. MoBA \citep{moba} scores
each block by the query's dot product with the block's mean key, per query head.

\textbf{Exact mass selection.} Deployed selectors estimate the attention mass that a block would receive
\citep{quest,nsa,seerattention}. The selection criterion that keeps the most attention mass for query $t$ at a given
budget is exact mass selection,
\begin{equation}
c_b^{\mathrm{exact}}=M_b=\sum_{g=1}^{G}\sum_{j\in \mathcal{B}_b}a^{(g)}_{t,j},
\label{eq:exact}
\end{equation}
which we use as an oracle in Section~\ref{sec:diagnosis}.

\section{Setting and Experimental Setup}
\label{sec:setup}

We study deployed sparse attention checkpoints at their shipped budget, through the model's own kernel and cache.
Every variant changes only which blocks are selected, except full attention in the flagged heads which reads every
block of those heads, and the per-layer budget which reads more blocks in the flagged layers.

We build on \infllm{} \citep{infllmv2}, whose selector combines fixed initial blocks, a local window
and top-$k$ blocks selected by \Eqref{eq:pool} \citep{streamingllm,nsa,moba}. Four checkpoints share this selector, the base model, the MiniCPM4-8B instruct model \citep{minicpm4}, \mcpm{} and a separate InfLLMv2-8B retraining of the same
backbone by \citet{dashattention}. Each has a full attention setting that differs by one configuration flag on
identical weights, and Table~\ref{tab:arch} gives their architecture and selector settings. \nosa{}-8B \citep{nosa}
adds a query-agnostic eviction head on top of \infllm{}. We also study other selector designs, the SeerAttention-R gates
on Qwen3-4B, 8B and 14B \citep{qwen3} and on DeepSeek-R1-Distill-Qwen-14B, the \seer{} gate on Llama-3.1-8B
\citep{llama3}, Dash-8B and NSA-8B, which \citet{dashattention} continued from MiniCPM4-8B, and MoBA fine-tunes of
Qwen3-8B and 4B released with Elastic Attention \citep{elasticattn}. As a contrast, we also run two token-level
selectors, DeepSeek-V3.2 \citep{dsa} and LongCat-Flash-Lite-Sparse \citep{longcatlsa}.

Our main task is \mk{3}, RULER's multi-key needle task \citep{ruler}, whose haystack is entirely
look-alike needles and on which selectors fail the most. We also report the other RULER tasks, HELMET \citep{helmet}
and five benchmarks of real data. Every comparison is paired on identical prompts from one seed of RULER's generator,
which also guided most design choices on the development models. The headline settings are repeated with six seeds,
and the five seeds held out from those choices carry the main results (Appendix~\ref{app:harness}). Every variant acts
through a hook on the selector's output that reproduces the released selector's selections exactly when the hook is
inactive (Appendix~\ref{app:setup}).

We call \emph{blind head} an attention head that attends to the needle while its selector misses the needle's block. A
\emph{flagged head} is a head that the probe selects. A \emph{fix} acts only in the flagged heads and takes one of
three forms, full attention, a larger budget in the flagged layers or the 4-bit scorer of Section~\ref{sec:method} at
the shipped budget. \emph{Decode tokens} are the generated tokens, \emph{question tokens} are the last prompt
tokens, which hold the question, and \emph{every token} adds the rest of the prompt. Unless stated otherwise, a fix acts at every token.
In the tables, \emph{Dense} is full attention in every head and \emph{Sparse} is the shipped selector.

\section{Diagnosing the Deficit}
\label{sec:diagnosis}

For each query token, layer and key-value head, exact attention mass selection computes the
true attention mass of every candidate block, summed over the query heads that share it. Summing per-head masses is the
head soft vote of \citet{tokenselect}. We use this selection as an instrument that diagnoses selector failures, not as a deployed path.

Formally, the instrument takes a set $\mathcal T$ of query tokens, from the prompt and from the generation, and a set
$\mathcal H$ of key-value heads. At a token $t$, it gives each head the selection
\begin{equation}
\mathcal{S}_t=\begin{cases}
\operatorname*{TopK}_{b\notin \mathcal{F}}\bigl(M_b,\,K\bigr) & \text{if } t\in\mathcal T \text{ and the head is in } \mathcal H,\\[2pt]
\mathcal{S}_t^{\mathrm{ship}} & \text{otherwise,}
\end{cases}
\label{eq:instrument}
\end{equation}
where $M_b$ is the exact mass of block $b$ at token $t$ from \Eqref{eq:exact} and $\mathcal{S}_t^{\mathrm{ship}}$ is the
shipped selector's choice. The forced set $\mathcal{F}$ and the budget $K$ stay those of the original model. Exact mass
keeps the most attention mass at the shipped budget, so the accuracy that \Eqref{eq:instrument} recovers over the
shipped selector estimates the loss caused by the choice of blocks at the tokens and heads in $\mathcal T$ and
$\mathcal H$. It is a mass-optimal reference, not a bound on accuracy.

\begin{table}[t]
\caption{\textbf{Where the multi-key loss is made.} Each row applies \Eqref{eq:instrument} at the tokens and heads its
label names and keeps the shipped selector elsewhere. Dense rows read every block without a selector. The mean over the
five seeds held out from every design choice, 200 prompts each.
In all tables, \textbf{bold} is the best number and \underline{underlined} the second best,
full attention excluded.}
\label{tab:diag}
\begin{center}
\footnotesize
\setlength{\tabcolsep}{3.5pt}
\fitwidth{%
\begin{tabular}{lccc}
\toprule
Selection & \infllm{}-8B, 32K & \mcpm{}, 32K & \mcpm{}, 64K \\
\midrule
Sparse & 63.1 & 82.7 & 54.1 \\
Exact mass at decode & 85.4 & 89.2 & 80.0 \\
Exact mass at decode and window & \textbf{95.3} & \textbf{100.0} & \textbf{97.5} \\
Exact mass at every token of flagged heads & \underline{94.7} & \textbf{100.0} & \underline{96.9} \\
Exact mass at decode tokens of flagged heads & 85.5 & \underline{89.3} & 80.2 \\
Dense at every token of flagged heads & 96.0 & 100.0 & 97.5 \\
Dense at every token of other heads & 65.2 & 91.5 & 61.8 \\
Dense & 97.8 & 100.0 & 97.1 \\
\bottomrule

\end{tabular}}
\end{center}
\end{table}

Within each checkpoint's native range, exact mass selection at the
shipped budget comes close to full attention (Table~\ref{tab:diag}).
The shipped budget can hold the needle's block, and the selector ranks it too low to be
kept. A larger budget in the three flagged layers alone, filled from the shipped ranking, recovers as much at 32K
(Table~\ref{tab:tradeoff}), because the needle's block sits within a few multiples of the budget in the shipped
ranking. On the retraining's smaller budget and beyond the native range, part of the loss is budget
(Table~\ref{tab:bounds}).

On \infllm{}-8B, exact mass at the decode tokens alone
recovers most of the deficit, and adding the question tokens recovers nearly all of it. On \mcpm{}, the question tokens
carry a larger share, and exact mass at the decode tokens alone stays far below full attention. A fix must therefore
also act inside the prefill, most of all on this checkpoint (Table~\ref{tab:diag}).

To find the heads that carry the loss, we run a needle calibration
probe on RULER \mk{3} prompts of a seed that no evaluation uses. The probe teacher-forces the answer's needle tokens and
reads two quantities in every selection unit $u$, a key-value head for the block selectors and a query head for MoBA.
The first is the needle attention that the selector forfeits, $F_u=\bigl(r_u(\mathcal E)-r_u(\mathcal S)\bigr)\,m_u$,
where $r_u(\mathcal E)-r_u(\mathcal S)$ is the needle recall that the shipped selector loses against exact mass at the
same budget and $m_u$ is the attention the unit puts on the needle block (\Eqref{eq:probe} in
Appendix~\ref{app:calib}). It explains why a head fails. The second chooses the heads to fix. Each unit receives a gate
$\alpha_u$ that blends, at every position $i$ where the selector acts, the output of each of its query heads $g$ over
the shipped selection with its output over every key,
\begin{equation}
\bm{o}^{(g)}_i(\bm\alpha)=\bm{o}^{(g),\mathrm{ship}}_i+\alpha_u\bigl(\bm{o}^{(g),\mathrm{full}}_i-\bm{o}^{(g),\mathrm{ship}}_i\bigr),
\label{eq:gate}
\end{equation}
as the gates of DuoAttention blend streaming and full attention \citep{duoattention}, so that $\bm\alpha=\bm 0$ is the
shipped model and $\bm\alpha=\bm 1$ is full attention in every unit. With $\mathcal F_x(\bm\alpha)$ the log-probability
of the needle tokens of calibration prompt $x$, the fix gain of unit $u$ is
\begin{equation}
\Gamma_u=\frac{1}{|\mathcal X_c|}\sum_{x\in\mathcal X_c}\,\Bigl|\int_0^1\frac{\partial\mathcal F_x}{\partial\alpha_u}\Big|_{\bm\alpha=\beta\bm 1}d\beta\,\Bigr|,
\qquad
\frac{\partial\mathcal F_x}{\partial\alpha_u}=\sum_{g\in u}\sum_{i}\nabla_{\bm{o}^{(g)}_i}\mathcal F_x^{\top}
\bigl(\bm{o}^{(g),\mathrm{full}}_i-\bm{o}^{(g),\mathrm{ship}}_i\bigr),
\label{eq:fixgain}
\end{equation}
the integrated gradient \citep{sundararajan2017ig} of the answer along the path from the shipped selector to full
attention, in absolute value per prompt and averaged over the calibration prompts $\mathcal X_c$. Summed over units, the
integrals add up to the gap $\mathcal F_x(\bm 1)-\mathcal F_x(\bm 0)$ when no selection switches along the path, so the
integral of a unit is its share of what full attention restores. The derivative at $\bm\alpha=\bm 0$ alone is the
attribution patching estimate of the fix \citep{nanda2023atp,kramar2024atpstar}, which misjudges heads that matter only
once others are fixed, and integrating along the path corrects this \citep{hanna2024faithfulness}
(Appendix~\ref{app:fixgain}). We evaluate the integral by the midpoint rule on $m=16$ points, each a forward and a
backward pass that scores every unit at once.

The flagged units are those above a threshold that is swept per checkpoint,
$\mathcal H(\tau)=\{u\mid\Gamma_u\ge\tau\}$, starting at the largest score and lowering $\tau$ by a factor of $\sqrt2$ per
step. Each step that adds units gives them full attention on the calibration prompts, and the rule keeps the first
threshold whose units close a share $\rho=0.9$ of the gap between the shipped selector and full attention in every unit.
This share is the faithfulness of circuit analysis \citep{wang2022ioi,marks2025sfc}, where a fixed cutoff near it is
common \citep{hanna2025formal}, and it is the level that \citet{prulong} ask of a long-context method relative to full
attention. The step gives a geometric grid, as in hyperparameter sweeps \citep{blake2024umup,everett2024scaling}. Neither
constant is tuned, and other shares and steps are in Table~\ref{tab:ruleablation}. How many heads carry the loss is
then a measurement on each checkpoint rather than a setting.

\textbf{The loss lives in three heads of the MiniCPM4-8B lineage.} On \infllm{}-8B, both quantities return the first
key-value head of layers 9, 16 and 22, far above every other head. The fix gain ranks the same three first on the
instruct checkpoint, the retraining and \mcpm{}, and its first-order version does on \nosa{}-8B, while the forfeited
attention places another late head third on \nosa{}-8B (Appendix~\ref{app:fixgain}). The three are flagged from each of five calibration seeds, at 16K and from
held-out JSON key-value data outside RULER. Exact mass in these heads alone recovers as much as exact mass in every
head, and full attention in these three heads recovers more of the loss than full attention in all the other heads
together (Table~\ref{tab:diag}). A scan of single
heads finds the same three, and three random heads change nothing (Table~\ref{tab:scan}, Appendix~\ref{app:harness}).
These heads attend to the needle whenever its block is selected, but on look-alike retrieval their selector does not
select it. As such, we call them \textbf{\textit{blind}} heads.

The retrieval score of \citet{retrievalhead} and the needle mass
alone rank heads whose selector already finds the needle above the blind heads (Table~\ref{tab:calibset}). The echo
score of RazorAttention, measured under its own protocol, ranks the three blind heads first on \infllm{}-8B but without a
break below them, and its rule of fixed fractions keeps a large share of the key-value heads.
The same heads fail under other selectors of the backbone. On Dash-8B and NSA-8B, the fix gain ranks the three heads
first, and the number of further heads that each selector needs differs (Section~\ref{sec:results}).

\begin{figure}[t]
\centering
\includegraphics[width=\linewidth]{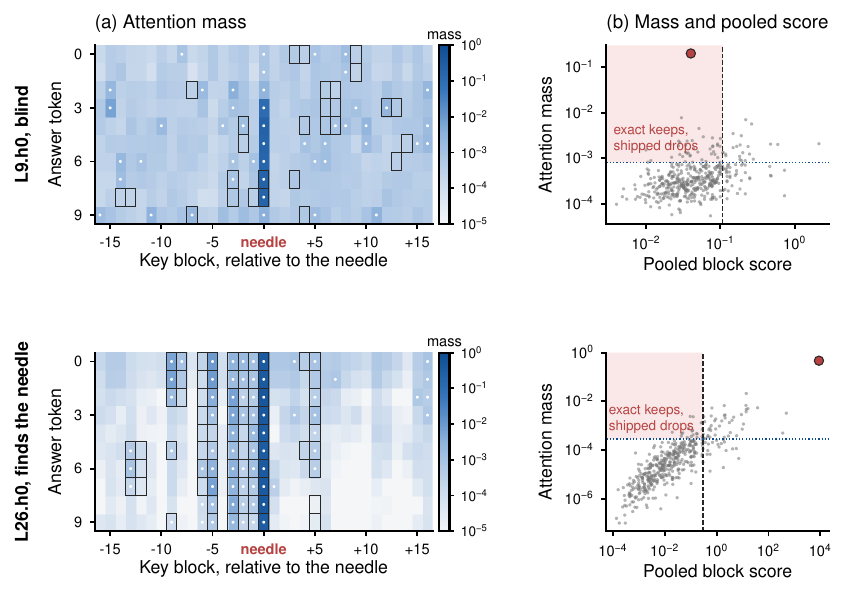}
\caption{\textbf{Inside a blind head.} One calibration prompt of \infllm{}-8B at 32K, ten teacher-forced answer tokens.
Top, the blind head L9.h0. Bottom, L26.h0, whose selector finds the needle. (a) Attention mass of the key blocks around
the needle, averaged over the query heads of the group. Outlined blocks are read by the shipped selector and dotted blocks
by exact mass selection. (b) At one answer token, the pooled score of every free block against its attention mass, the
needle block in red, with the shipped cutoff dashed and the exact mass cutoff dotted. The blind head attends to the
needle, but the pooled score leaves the needle block below the shipped cutoff, while exact mass keeps it.}
\label{fig:inside}
\end{figure}

\begin{figure}[t]
\centering
\includegraphics[width=\linewidth]{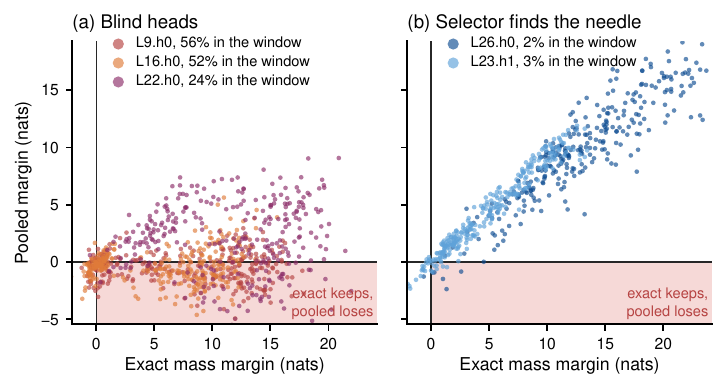}
\caption{\textbf{Pooling erases the needle's lead} on every probed token of the calibration prompts, \infllm{}-8B at
32K, in the query head of each group with the most needle mass. The horizontal axis is the needle block's lead over the
block at the selection cutoff under exact mass. The vertical axis is its lead under the pooled score. Both axes are in nats, the units
of the natural logarithm \citep{coverthomas}, so that each sign reproduces whether that selection keeps the needle. Where the selector finds the needle, the two leads
move together. In the blind heads, the pooled lead stays near zero whatever the exact lead, and a quarter to more than
half of the tokens fall in the shaded window of Proposition~\ref{prop:pool}, kept by exact mass and lost by pooling.}
\label{fig:window}
\end{figure}

\textbf{Pooling loses the needle.} In the blind heads of layers 9 and 16, only a few of the group's query heads put
appreciable attention on the needle block. Scored alone, each of them usually finds the needle block under its own
token-level attention but misses it under its softmax over the mean-pooled keys (Table~\ref{tab:probed}). Figure~\ref{fig:inside} shows one prompt. A two-block
model organizes these measurements. Let a block hold one needle key whose logit exceeds its neighbors' by $\Delta$, and
let a competitor block have the logit of every key elevated by $\delta$. Exact mass ranks the needle block first
whenever $\Delta>\delta+\log L$, while any score that increases with the mean logit over a window of $w$ keys ranks the
competitor first whenever $\Delta<w\delta$, so every needle in between is lost to pooling
(Proposition~\ref{prop:pool}, Figure~\ref{fig:window}). The interval widens with the number of competitors, as the deficit does with length
(Figure~\ref{fig:lensweep}), and closes when no competitor resembles the needle, which is why the calibration haystack
decides which heads are flagged. Dilution of this kind accounts for most of the shipped misses in all three blind
heads, and the rest are lost to competitor blocks that outvote the needle (Table~\ref{tab:anatomy}). Quantizing the keys moves each block score by at most its largest logit error
(Proposition~\ref{prop:quant}), and on the probed tokens the needle block keeps its rank with keys quantized to a few
bits and loses it with keys pooled.

\section{Training-Free Fixes for the Flagged Heads}
\label{sec:method}

\begin{figure}[t]
\centering
\includegraphics[width=0.9\linewidth]{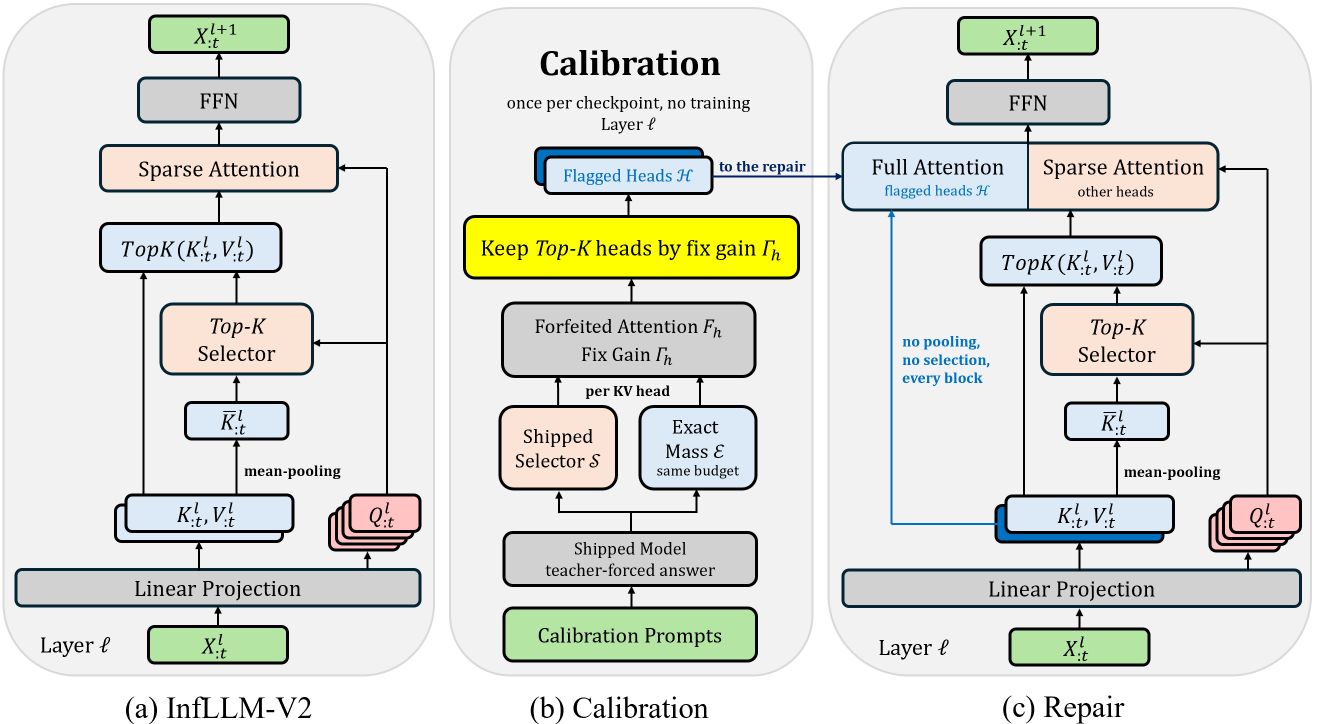}
\caption{\textbf{Method overview.} (a) \infllm{} scores each block from mean-pooled keys, gathers the top-$k$ blocks and
attends to them in every head. (b) A needle probe, run once per checkpoint, compares the shipped selection with exact
attention mass at the same budget, which explains the failure, and keeps the heads with the largest gain of fixing
their selection. (c) In the flagged heads,
full attention reads every block, and every other head keeps the shipped selector. A 4-bit scorer that keeps the shipped
budget can replace full attention in those heads.}
\label{fig:arch}
\end{figure}

Once the flagged heads are known, the simplest fix gives them full attention (Figure~\ref{fig:arch}), as DuoAttention and RazorAttention do for
retrieval heads. It reads every key and value of those heads at every step, which costs little for a single sequence and
grows with batch and context. The diagnosis points to two cheaper forms. A larger budget in the flagged layers needs no
new code. A scorer that keeps token resolution at reduced precision keeps the shipped budget and changes only the
top-$k$ path in the flagged heads, and it also tests the account above, since it restores resolution and drops
precision.

\textbf{Calibration.} The probe and the threshold sweep run once per checkpoint on calibration prompts at 32K, or 16K
for Dash-8B and NSA-8B. The process trains nothing and takes hours on one GPU, most of it the backward passes of the integrated probe
(Table~\ref{tab:calcost}). Head-level calibration on a small synthetic retrieval set is common
\citep{duoattention,retrievalhead,minference}. The rule sets the number of flagged heads on every checkpoint outside the lineage. On the \infllm{} lineage, the rule keeps the three blind heads with at most one head more or fewer, and the lineage
keeps the three of the diagnosis (Table~\ref{tab:lineage}). A probe needs lookalike
competitors, so a single needle among repeated filler sentences flags no head at all (Table~\ref{tab:calibset}).

\textbf{Side cache.} The scorer keeps a 4-bit copy of each flagged head's keys, used only for scoring. Each channel $c$
is standardized by its mean $\mu_c$ and standard deviation $\sigma_c$ and quantized uniformly with the step $\Delta_b$
that is optimal for a Gaussian source at $b$ bits \citep{max1960},
\begin{equation}
\bm{\hat{k}}_{j,c}=\bm{\mu}_c+\sigma_c\Delta_b\,z_{j,c},\qquad
z_{j,c}=\operatorname{clamp}\Bigl(\Bigl\lfloor\frac{\bm{k}_{j,c}-\bm{\mu}_c}{\sigma_c\Delta_b}\Bigr\rfloor+\tfrac12,\;-2^{b-1}+\tfrac12,\;2^{b-1}-\tfrac12\Bigr).
\label{eq:quant}
\end{equation}
Here, $z_{j,c}$ is the $b$-bit code stored for $k_{j,c}$ and $\hat k_{j,c}$ is the reconstructed key used to score
blocks. Standardizing each channel lets one step fit channels whose spreads differ widely, as in per-channel key caches
\citep{kivi}.
The clamp can only change which blocks score highest, since attention reads the full-precision keys and always reads the
forced blocks, the attention sink \citep{streamingllm} among them. We use $b=4$, at which the scorer keeps the needle
block as well as exact mass on the probed tokens (Table~\ref{tab:noise}). When the scorer acts at every prompt token,
$\mu_c$ and $\sigma_c$ come from a fixed prefix of the prompt, so the code of a key never depends on later tokens. The
copy adds under one percent to the key-value cache.

\textbf{Selection.} At every token of a flagged head, each query head of the group takes a softmax over the low-bit keys
up to its own position. The distributions are summed within each block, and the forced blocks and the best-scoring free
blocks fill the shipped budget. Every other head keeps the shipped selection, and the kernel and the cache are unchanged.
The scorer is the instrument of \Eqref{eq:instrument} with $\mathcal H$ the flagged heads and the exact mass $M_b$
replaced by its value on the 4-bit keys,
\begin{equation}
\hat M_b=\sum_{g=1}^{G}\sum_{j\in \mathcal{B}_b}\operatorname{softmax}_j\!\left(\frac{\mathbf{q}_t^{(g)\top}\hat{\mathbf{k}}_j}{\sqrt d}\right),
\label{eq:repair}
\end{equation}
where $\hat{\mathbf{k}}_j$ is the 4-bit copy of key $j$. Its three variants act at the decode tokens, also at the
question tokens or at every token.

\section{Results}
\label{sec:results}

\begin{table}[t]
\caption{\textbf{Multi-key retrieval (\mk{3}) on the five seeds held out from every design choice}, mean over five seeds. The first seed guided the design
choices. ``Full Attention, Flagged Heads'' gives the flagged heads full attention at every token.
``Decode'', ``+ Question'' and ``Every Token'' are the 4-bit scorer in the flagged heads at the decode tokens, also at
the question tokens and at every token.}
\label{tab:seeds}
\begin{center}
\footnotesize
\setlength{\tabcolsep}{3.5pt}
\fitwidth{%
\begin{tabular}{llcccccc}
\toprule
 & & & & & \multicolumn{3}{c}{4-bit Scorer, Flagged Heads} \\
\cmidrule(lr){6-8}
Model & Context & Dense & Sparse & \begin{tabular}[b]{@{}c@{}}Full Attention,\\Flagged Heads\end{tabular} & Decode & + Question & Every Token \\
\midrule
\infllm{}-8B & 32K & 97.8 & 63.1 & \textbf{96.0} & 85.6 & \underline{95.1} & 94.7 \\
\mcpm{} & 32K & 100.0 & 82.7 & \textbf{100.0} & 89.1 & \underline{99.9} & \textbf{100.0} \\
\mcpm{} & 64K & 97.1 & 54.1 & \textbf{97.5} & 79.7 & \underline{97.2} & \textbf{97.5} \\
\bottomrule
 
\end{tabular}}
\end{center}
\end{table}

\textbf{Every fix recovers most of the multi-key loss.} On the held-out seeds, full attention in the flagged heads and
the scorer at every token come within a few points of full attention on \infllm{}-8B and match it on \mcpm{}, and the
question-token variant recovers about as much (Table~\ref{tab:seeds}). On the retraining and on \nosa{}-8B, whose loss
is made across the prefill, the scorer at every token closes more of the gap than the question-token variant, and full
attention in the flagged heads closes more still, the most on the retraining whose smaller budget binds
(Table~\ref{tab:bounds}). At 64K, twice the base model's native range, no fix reaches full attention, and each beats the
shipped selector given several times as many blocks per token.

\begin{figure}[t]
\centering
\includegraphics[width=\linewidth]{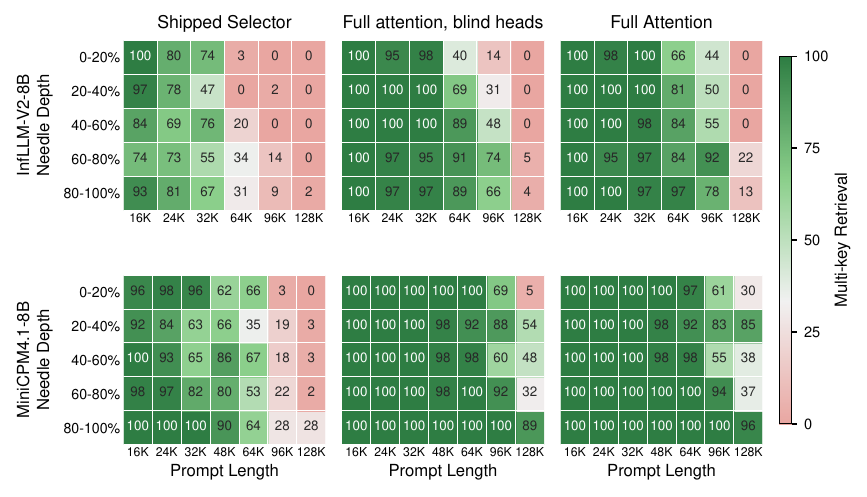}
\caption{\textbf{The loss by needle depth.} \mk{3} accuracy in five bands of the needle's position in the prompt, at
prompt lengths from 16K to 128K, for the shipped selector, full attention in the blind heads and full attention. The
native range ends at 32K for \infllm{}-8B and at 64K for \mcpm{}. About 40 prompts per cell, from the 200 prompts of the
first seed at each length.}
\label{fig:depth}
\end{figure}

Figure~\ref{fig:depth} splits \mk{3} by the needle's depth in the prompt at prompt
lengths from 16K to 128K. Within each model's native range, the shipped selector loses needles at every depth, and full
attention in the blind heads brings every band close to full attention. Beyond the native range, the shipped selector
loses the earliest needles first, and full attention itself starts to lose them too (Figure~\ref{fig:lensweep} gives the
mean over depths).

\begin{table}[t]
\caption{\textbf{What each fix gains and costs} on \infllm{}-8B at 32K. \mk{3} and common word extraction (CWE) as mean over five seeds.
PopQA, Natural Questions (NQ) and HotpotQA are HELMET's retrieval-augmented question answering at its default size,
600 examples each from 100 questions for PopQA and NQ and from 200 for HotpotQA, every arm on H100. Bytes are read per cached token and decode step beyond the blocks every sparse method reads, as in
Table~\ref{tab:compare}, at 32K for the larger budget, whose extra blocks are a fixed count per step. The two
rows marked 16 CWE heads run the scorer in the 16 heads calibrated on common word extraction's own prompts.}
\label{tab:tradeoff}
\begin{center}
\footnotesize
\setlength{\tabcolsep}{3pt}
\fitwidth{%
\begin{tabular}{lcccccc}
\toprule
Fix & \mk{3} & CWE & PopQA & NQ & HotpotQA & Bytes per Cached Token \\
\midrule
Shipped selector & 63.1 & 45.5 & \textbf{41.2} & 44.7 & 41.5 & 1,024 \\
Full attention, flagged heads & 96.0 & 33.9 & 38.0 & 44.8 & 43.5 & 2,512 \\
4x budget, flagged layers & \textbf{97.2} & 36.6 & 37.7 & \underline{44.8} & \textbf{43.7} & 2,140 \\
Scorer, every token & 94.7 & 40.0 & 37.8 & 43.7 & \underline{42.3} & 1,168 \\
Scorer, + question & \underline{95.1} & 45.7 & 37.2 & 44.2 & 41.2 & 1,168 \\
Scorer, decode & 85.6 & 45.6 & \underline{41.0} & \textbf{45.2} & 42.2 & 1,168 \\
Scorer, decode, 16 CWE heads & 85.1 & \underline{51.4} & 40.0 & 43.3 & 42.2 & 1,792 \\
Scorer, + question, 16 CWE heads & 94.2 & \textbf{52.7} & 37.2 & 41.5 & 41.7 & 1,792 \\
Dense & 97.8 & 51.6 & 39.5 & 42.5 & 42.5 & 32,768 \\
\bottomrule
 
\end{tabular}}
\end{center}
\end{table}

Every fix that recovers nearly all of the multi-key loss costs accuracy on common word
extraction, which aggregates over the whole prompt, or on PopQA (Table~\ref{tab:tradeoff}). Full attention in the flagged
heads pays the most on common word extraction, and the larger budget nearly as much. On \mcpm{}, full attention in the
flagged heads and the larger budget also inherit part of the variable tracking loss of full attention (Table~\ref{tab:aggregation}). The scorer at every token
pays less there. The question-token variant pays nothing there but loses on PopQA, as does the larger budget. Only the
scorer at the decode tokens is harmless on every task we measured, and it recovers less of the multi-key loss. On NQ and HotpotQA, no fix in the three flagged heads costs more than a point.

On PopQA, the shipped selector is ahead of full attention itself. Giving a head the needle
helps when one token decides the answer. When the answer aggregates over many tokens, the same head's shipped selection
may already be the better summary.

The common word loss itself sits in other heads. We rank the heads by the
attention mass that their shipped selection forfeits on the task's own prompts, which needs no labels, and take the 16
highest-ranked heads, the three blind heads among them. The scorer at the decode tokens of those 16 heads recovers common
word extraction and keeps the multi-key gain, and with the question tokens it recovers nearly all of both at a larger
cost on question answering (Table~\ref{tab:tradeoff}). More heads with full attention reach a similar end at
several times the bytes.

\begin{table}[t]
\caption{\textbf{Choosing a fix for an unknown workload} on \infllm{}-8B at 32K. The fixes of
Table~\ref{tab:tradeoff} on
17 task types of the first seed, the 12 RULER tasks, HELMET's PopQA, Natural Questions, HotpotQA and JSON key-value
retrieval, and LongFuncEval. ``Worst Change'' is the largest loss against the shipped selector on any one task, with that
task, and ``Mean Change'' averages the 17 tasks. ``Beats Shipped'' is the share of 10,000 random mixtures of the tasks in
which the fix scores above the shipped selector, with weights drawn uniformly over the tasks and, after the slash, over
the three benchmarks and then their tasks.}
\label{tab:mixcrit}
\begin{center}
\footnotesize
\begin{tabular}{lccc}
\toprule
Fix & Worst Change & Mean Change & Beats Shipped (\%) \\
\midrule
Full attention, flagged heads & $-12.5$ (Common Words) & $+3.7$ & 97.3 / 98.8 \\
4x budget, flagged layers & $-9.6$ (Common Words) & \underline{$+3.8$} & 98.5 / 99.0 \\
Scorer, every token & $-5.5$ (Common Words) & $+3.6$ & 98.9 / 98.3 \\
Scorer, + question & $-4.0$ (PopQA) & $+3.7$ & \underline{99.8 / 98.0} \\
Scorer, decode & \textbf{\boldmath$-0.9$ (Multi-value)} & $+2.5$ & \textbf{100.0 / 100.0} \\
Scorer, decode, 16 CWE heads & \underline{$-1.7$ (Frequent Words)} & $+2.7$ & 99.8 / 92.0 \\
Scorer, + question, 16 CWE heads & $-4.0$ (PopQA) & \textbf{\boldmath$+3.9$} & 99.6 / 97.0 \\
\bottomrule

\end{tabular}
\end{center}
\end{table}

\textbf{Choosing a fix for an unknown workload.} The accuracy of a workload is the weighted mean of its tasks'
accuracies, so the worst workload for a fix is a single task. Table~\ref{tab:mixcrit} sets the fixes against each other
on 17 task types. Only the scorer at the decode tokens of the three flagged heads never loses more than a point on any
task, and it beats the shipped selector on nearly every random mixture of the tasks. Every fix that rewrites prompt tokens gains
more on average and pays for it on one kind of task: the question-token variants on open-domain question answering, and
full attention, the larger budget and the every-token scorer on common word extraction. The scorer at the decode tokens
is therefore the safe default when the workload is unknown. The question tokens or heads calibrated on the task help
where the workload is known to lean on retrieval or aggregation and natural-text question answering is not at stake
(Table~\ref{tab:tradeoff}).

On five benchmarks of real data with the lineage's checkpoints, the one deficit is on identifier
retrieval, where the fixes recover part of it (Table~\ref{tab:real}). Section~\ref{sec:selectors} adds a benchmark on
which the SeerAttention-R gate loses accuracy.

\begin{table}[t]
\caption{\textbf{Against published selectors} on the two checkpoints that \citet{sparda} evaluate, \mk{3} and the mean
of RULER's 12 tasks at the shipped budget. Quest scores a block by the bound from the per-channel extremes of its keys,
Double Sparsity and Loki by the group's attention on 16 label channels and on 32 principal directions chosen from each
prompt's own keys, all three in every head at the decode and question tokens, and SnapKV \citep{snapkv} selects at
decode the blocks that the last 32 prompt tokens attend to most. ``Loki, flagged heads'' acts at every token of the
flagged heads at the scorer's bytes. Bytes per cached token is what a method reads per cached token and decode step
beyond the blocks every sparse method reads, and the whole cache for Dense.}
\label{tab:compare}
\begin{center}
\footnotesize
\setlength{\tabcolsep}{4pt}
\begin{tabular}{lccccc}
\toprule
 & \multicolumn{2}{c}{\mcpm{}, 64K} & \multicolumn{2}{c}{\nosa{}-8B, 32K} & Bytes per \\
\cmidrule(lr){2-3}\cmidrule(lr){4-5}
 & \mk{3} & RULER & \mk{3} & RULER & Cached Token \\
\midrule
Dense & 97.0 & 86.9 & 94.0 & 89.4 & 32,768 \\
Sparse & 55.5 & 80.2 & 18.5 & 70.6 & 1,024 \\
Quest & 22.0 & 73.8 & 18.5 & 74.2 & 512 \\
Double Sparsity & 34.0 & 72.6 & 29.5 & 72.5 & 2,048 \\
Loki & 96.0 & \textbf{88.0} & 83.0 & \textbf{86.7} & 4,096 \\
SnapKV & 68.0 & 81.7 & 24.5 & 73.9 & 0 \\
\rowcolor{oursbg} Full attention, flagged heads & 97.5 & 86.0 & 94.0 & 85.8 & 2,512 \\
\rowcolor{oursbg} Loki, flagged heads & \textbf{97.5} & \underline{87.4} & \underline{86.0} & 84.7 & 1,168 \\
\rowcolor{oursbg} Scorer, + question & 96.5 & 86.8 & 81.0 & 84.8 & 1,168 \\
\rowcolor{oursbg} Scorer, every token & \underline{97.0} & 87.2 & \textbf{89.0} & \underline{85.2} & 1,168 \\
\bottomrule

\end{tabular}
\end{center}
\end{table}

\textbf{Token resolution matters more than precision.} At the same budget and in every head, Quest's bound and Double
Sparsity's label channels fall below the shipped selector on \mk{3} of \mcpm{}, and SnapKV recovers part of the loss
(Table~\ref{tab:compare}). Loki's principal directions \citep{loki} in every head match the question-token scorer on
\mk{3} of \mcpm{} and do better on the 12-task means and on \nosa{}-8B, at several times its bytes. In the flagged heads at matched bytes, keys of a few bits
\citep{sale,ffd} and Loki's directions reach exact mass, while Quest's bound at block granularity and mean-pooled keys
under the same criterion fall below the shipped selector (Appendix~\ref{app:setup}). At the shipped budget, what
matters is token resolution in the right heads. Every-head summaries at similar bytes are at least as good on the wider
suite, and localizing the fix to a few heads mainly saves side-cache memory.

\section{One Rule Across Selector Designs}
\label{sec:selectors}

The three heads of the lineage come from the diagnosis. On the other selector designs, the rule of
Section~\ref{sec:diagnosis} sets the heads, with the same probe, share and step on every checkpoint. The
SeerAttention-R gates \citep{seerattentionr} on Qwen3-8B, Qwen3-4B, Qwen3-14B and DeepSeek-R1-Distill-Qwen-14B, a
reasoning-distilled model of another backbone, select blocks at decode in every key-value head. Dash-8B
\citep{dashattention} and NSA-8B \citep{nsa} select under other block selectors on the MiniCPM4-8B backbone. The MoBA
fine-tunes of Qwen3-8B and 4B \citep{moba} score long chunks by their mean key in every query head, and since they barely
retrieve on \mk{3} even with full attention, they use RULER's single-needle task. On each checkpoint, the probe ranks
the units on calibration prompts, the threshold sweep sets their number and every arm then runs on the five evaluation
seeds. RazorAttention's echo heads \citep{razorattention} and random heads of the same count are the comparators.

\begin{table}[t]
\caption{\textbf{The rule on other selector designs.} The flagged heads are those above the threshold that closes
$\rho=0.9$ of the gap on calibration prompts, key-value heads for the block selectors and query heads for MoBA. ``Full Attention'' and
``4-bit Scorer'' are the two fixes in the flagged heads. The scorer acts at each release's budget, at the decode tokens on
Dash-8B and at every token on MoBA. ``Echo'' and ``Random'' give full attention to as many heads by RazorAttention's echo
score and at random. $^{\ddagger}$NSA-8B ships no dense path, so its dense column selects every
block.}
\label{tab:rule}
\begin{center}
\footnotesize
\setlength{\tabcolsep}{3pt}
\fitwidth{%
\begin{tabular}{lllcc>{\columncolor{oursbg}}cccc}
\toprule
 & & & & & \multicolumn{2}{c}{Flagged Heads} & \multicolumn{2}{c}{Same Count} \\
\cmidrule(lr){6-7}\cmidrule(lr){8-9}
Model & Selector & Heads & Dense & Sparse & Full Attention & 4-bit Scorer & Echo & Random \\
\midrule
Qwen3-8B & SeerAttention-R & 18 of 288 & 98.7 & 29.8 & \textbf{97.7} & \underline{97.3} & 69.4 & 31.4 \\
Qwen3-4B & SeerAttention-R & 9 of 288 & 97.5 & 16.6 & \underline{87.8} & \textbf{88.9} & 56.3 & 17.5 \\
Qwen3-14B & SeerAttention-R & 5 of 320 & 99.7 & 54.5 & \underline{96.9} & \textbf{98.1} & 64.1 & 54.6 \\
R1-Distill-Qwen-14B & SeerAttention-R & 22 of 384 & 85.6 & 17.9 & \textbf{79.1} & \underline{73.9} & 42.9 & 21.2 \\
Dash-8B & DashAttention & 13 of 64 & 99.4 & 66.6 & \textbf{98.0} & \underline{92.0} & 80.9 & 91.0 \\
NSA-8B$^{\ddagger}$ & NSA & 4 of 64 & 98.1 & 14.1 & \textbf{95.2} & \underline{94.5} & 68.2 & 13.5 \\
MoBA, Qwen3-8B & MoBA & 513 of 1152 & 94.4 & 21.1 & \textbf{89.6} & 75.8 & \underline{85.9} & 50.5 \\
MoBA, Qwen3-4B & MoBA & 433 of 1152 & 85.7 & 15.2 & \underline{80.0} & 74.1 & \textbf{87.6} & 25.2 \\
\bottomrule
 
\end{tabular}}
\end{center}
\end{table}

\begin{figure}[t]
\centering
\includegraphics[width=\linewidth]{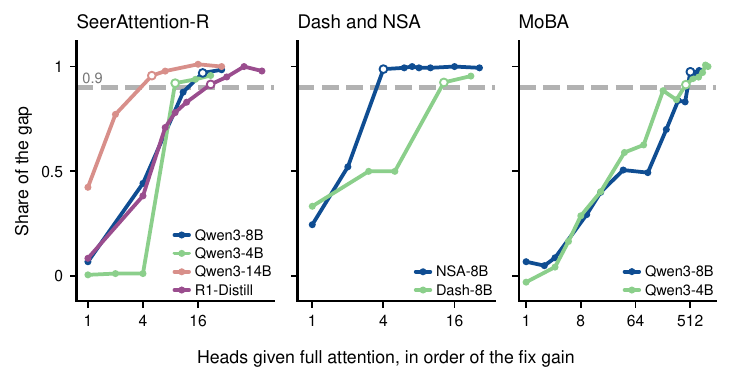}
\caption{\textbf{How many heads carry the loss.} The share of the gap between the shipped selector and full attention
in every head that full attention in the top heads by fix gain closes on calibration prompts, at each step of the
threshold sweep. The dashed line is the share $\rho=0.9$, and the marked point is the count the rule keeps.}
\label{fig:rule_curves}
\end{figure}

\textbf{A few heads carry the loss under three selectors.} On the SeerAttention-R gates, Dash-8B and NSA-8B, full
attention in the flagged heads recovers most of the loss, close to full attention in every head on all but the Qwen3-4B
gate (Table~\ref{tab:rule}). As many heads by echo score recover part of it, and random heads recover little, except on
Dash-8B, where so large a random set often includes a blind head. The 4-bit scorer in the same heads recovers as much as
full attention on the Qwen3 gates and on NSA-8B, so the fix that keeps the shipped budget suffices there, and less on
R1-Distill and on Dash-8B, where it acts at the decode tokens only.

On NSA-8B, the flagged heads are the three blind heads
of the lineage and one other. On Dash-8B, whose shipped selector loses less, the three rank first and the rule adds
further heads. Which heads fail is thus shared across the selectors of one backbone, and how many further heads fail
depends on the selector.

\textbf{On MoBA, the loss is not localized.} On both MoBA fine-tunes, the rule flags a large share of the query heads, and
as many heads by echo score recover nearly as much on one and more on the other (Table~\ref{tab:rule}). The scorer in the
flagged heads recovers less than full attention there. The sweep rises slowly with the number
of heads (Figure~\ref{fig:rule_curves}). MoBA scores long chunks by their mean key in every query head and spreads its
loss over many of them, so no small set of heads carries it, and at that scale the probe's ranking adds little over a
retrieval score of the dense model.

\textbf{How few heads carry the loss depends on the selector.} The threshold sweep answers this per checkpoint
(Figure~\ref{fig:rule_curves}). A small share of the heads closes the gap under the SeerAttention-R gates and NSA, a
larger one under Dash, and a large one under MoBA. The rule thus measures, for a given selector, how far the
localization of the MiniCPM4-8B lineage holds.

\begin{table}[t]
\caption{\textbf{OpenAI MRCR on the SeerAttention-R gates}, by its official metric on all 593 prompts of its bins from 8K
to 32K. The flagged heads are those of Table~\ref{tab:rule}, chosen on synthetic multi-key prompts. ``Random'' is the mean
over three random sets of as many heads, none of them flagged.}
\label{tab:mrcr}
\begin{center}
\footnotesize
\begin{tabular}{llcc>{\columncolor{oursbg}}ccc}
\toprule
 & & & & \multicolumn{2}{c}{Flagged Heads} & Same Count \\
\cmidrule(lr){5-6}\cmidrule(lr){7-7}
Model & Heads & Dense & Sparse & Full Attention & 4-bit Scorer & Random \\
\midrule
Qwen3-8B & 18 of 288 & 36.0 & 29.2 & \textbf{33.2} & \textbf{33.2} & \underline{29.3} \\
Qwen3-14B & 5 of 320 & 35.2 & 32.7 & 32.7 & \textbf{33.3} & \underline{33.0} \\
\bottomrule

\end{tabular}
\end{center}
\end{table}

\textbf{The flagged heads carry over to a real benchmark.} The rule flags heads on synthetic multi-key prompts.
Table~\ref{tab:mrcr} runs OpenAI MRCR \citep{mrcr}, where the model must return one of several look-alike pieces of text
from a long conversation, on the SeerAttention-R gates over Qwen3-8B and Qwen3-14B, the strongest models with a dense
setting that the fixes apply to. On Qwen3-8B, the gate loses several points, and full attention or the 4-bit scorer in the
flagged heads recovers more than half of the loss, while random heads of the same count recover almost none. No MRCR prompt was
used to choose the heads. On Qwen3-14B, the gate loses less, and neither the flagged heads nor random heads recover it,
so the flagged heads do not explain that smaller loss.

\textbf{Token-level selectors show no such loss.} The account of Section~\ref{sec:diagnosis} ties the loss to block
scores computed from pooled keys. A token-level selector scores each key on its own and pools nothing, the case $w=1$ of
Proposition~\ref{prop:pool}, for which the blind window is empty. Its selection is also shared by the heads of a layer, so
no head can be blind in this sense. In both models we run, the indexer picks 2,048 tokens for each query. On RULER's \mk{3} per length, DeepSeek-V3.2 \citep{dsa} with its shipped DSA selection scores 100.0 at 32K, 64K and
128K, against 100.0, 100.0 and 98.0 for the same weights with every token selected. LongCat-Flash-Lite-Sparse \citep{longcatlsa} scores 99.4 at 32K and 99.3 at 64K, against 99.5 and 99.3 with every token
selected.

\begin{figure}[t]
\centering
\includegraphics[width=\linewidth]{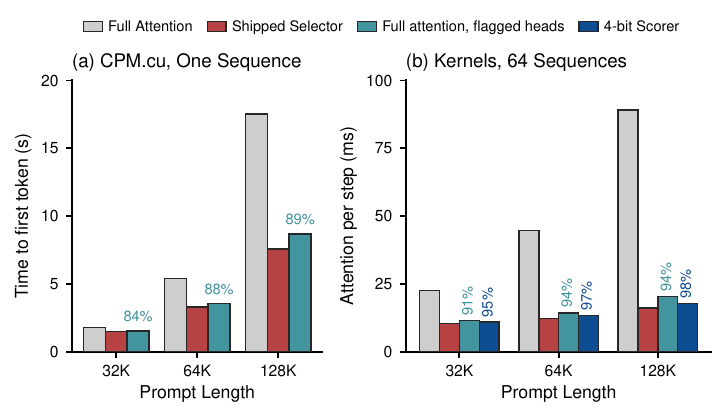}
\caption{\textbf{Latency against the dense model} on one H100. (a) TTFT of \infllm{}-8B at batch size
one in CPM.cu, the inference engine that OpenBMB ships for these checkpoints. (b) Attention time of one decode step of 64
sequences over all 32 layers in the release's kernels. The label above each bar of a fix is the share of the sparse
model's saving that it keeps.}
\label{fig:latency}
\end{figure}

\textbf{Cost.}\label{sec:cost} We time the fixes against the shipped selector and against the dense model
(Figure~\ref{fig:latency}). In CPM.cu, the inference engine that OpenBMB ships for these checkpoints, full attention in
the flagged heads keeps most of the TTFT that sparse attention saves, at every length. One sequence on
an H100 is bound by reading the weights, so sparse and dense decoding stay close per output token there, and the fix adds
a few percent. Sparse attention pays at decode once many long sequences share the step. There, both fixes keep most of
its saving in the release's attention kernels, and the scorer adds less than full attention in the flagged heads, as the
bytes that each reads suggest. The 4-bit scorer keeps the shipped budget, and its deployable path scans a packed side
cache with Triton kernels \citep{triton}. At the prefill, it costs more than full attention in the
flagged heads when it rewrites every prompt token, while the decode-token variant leaves the prefill's selection as shipped
(Table~\ref{tab:latency}). Full attention in the flagged heads is thus our default.

\textbf{Limitations.}\label{sec:limits} The three blind heads belong to one lineage. Its checkpoints share a backbone and
may inherit them \citep{retrievalhead}, and we have no second independently pretrained checkpoint with this selector and a
working dense mode. Other selectors need more heads, and MoBA spreads the loss over many. Calibration needs look-alike
competitors, and a head set calibrated on one task is tested only on the tasks we report. Latency
is measured in one production engine at batch size one, in the release's kernels for many sequences and in one serving
stack. The larger per-layer budget is measured on one seed at 64K and on no
real-text task beyond HELMET. The two token-level models are each measured on one multi-key task, where both score near
the ceiling.

\section{Conclusion}
\label{sec:conclusion}

Exact attention mass at the shipped budget shows that block selectors in deployed sparse attention models lose
retrieval by ranking the needle too low, and a calibration probe locates the heads where this happens. In the
MiniCPM4-8B lineage, the loss sits in three blind heads, whose mean-pooled key summary dilutes a single decisive token.
Fixing those heads recovers most of the loss without training. How concentrated the loss is depends on the selector. One rule, a threshold on the fix gain swept per
checkpoint, finds a few heads under the SeerAttention-R gates and NSA, a larger set under DashAttention led by the same
three blind heads, and no small set under MoBA. In the producers' engine, the default fix keeps most of the TTFT that sparse attention saves, a scorer at the decode tokens is safe across the tasks we measured and
token-level selectors show no such loss.

\subsection*{Reproducibility statement}
Every RULER and JSON key-value cell of the main text uses 200 prompts per seed unless stated otherwise. HELMET's real-text sets use
HELMET's default sizes. The real-data benchmarks use their own instance counts, stated per cell. Every result file
carries per-sample scores, its run settings, the data hash, and the counts of hook calls, rewritten selections and
errors, which a run must match to complete. Every variant is verified by kernel-call counting and by a fidelity check
that reproduces the released selections exactly (Appendix~\ref{app:harness}), and every fix-gain probe checks its
recomputed attention against the kernel's output or against the release's own decode. The code release holds the calibration probe with the threshold rule, the fixes as hooks
on the \infllm{} release and the flagged heads of six checkpoints, at
\url{https://github.com/zhangwenhe1007/blind-attention-heads}.

\bibliography{refs}

\begin{thebibliography}{79}
\providecommand{\natexlab}[1]{#1}
\providecommand{\url}[1]{\texttt{#1}}
\expandafter\ifx\csname urlstyle\endcsname\relax
  \providecommand{\doi}[1]{doi: #1}\else
  \providecommand{\doi}{doi: \begingroup \urlstyle{rm}\Url}\fi

\bibitem[Ahmad \& Yun(2026)Ahmad and Yun]{spotattention}
Huzama Ahmad and Se-Young Yun.
\newblock {SpotAttention}: Plug-in block-sparse routing for pretrained long-context transformers.
\newblock \emph{arXiv preprint arXiv:2606.22874}, 2026.

\bibitem[Ainslie et~al.(2023)Ainslie, Lee-Thorp, de~Jong, Zemlyanskiy, Lebr{\'o}n, and Sanghai]{gqa}
Joshua Ainslie, James Lee-Thorp, Michiel de~Jong, Yury Zemlyanskiy, Federico Lebr{\'o}n, and Sumit Sanghai.
\newblock {GQA}: Training generalized multi-query transformer models from multi-head checkpoints.
\newblock In \emph{Proceedings of the 2023 Conference on Empirical Methods in Natural Language Processing}, pp.\  4895--4901, 2023.

\bibitem[{Anthropic}(2026)]{claudeopus55}
{Anthropic}.
\newblock Introducing {Claude Opus 5.5}.
\newblock \url{https://www.anthropic.com/claude-opus-5-5}, September 2026.

\bibitem[Bertsch et~al.(2025)Bertsch, Pratapa, Mitamura, Neubig, and Gormley]{oolong}
Amanda Bertsch, Adithya Pratapa, Teruko Mitamura, Graham Neubig, and Matthew~R. Gormley.
\newblock Oolong: Evaluating long context reasoning and aggregation capabilities.
\newblock \emph{arXiv preprint arXiv:2511.02817}, 2025.

\bibitem[Bhaskar et~al.(2025)Bhaskar, Wettig, Gao, Dong, and Chen]{prulong}
Adithya Bhaskar, Alexander Wettig, Tianyu Gao, Yihe Dong, and Danqi Chen.
\newblock Cache me if you can: How many {KVs} do you need for effective long-context {LMs}?
\newblock \emph{arXiv preprint arXiv:2506.17121}, 2025.

\bibitem[Blake et~al.(2024)Blake, Eichenberg, Dean, Balles, Prince, Deiseroth, Cruz-Salinas, Luschi, Weinbach, and Orr]{blake2024umup}
Charlie Blake, Constantin Eichenberg, Josef Dean, Lukas Balles, Luke~Y. Prince, Bj{\"o}rn Deiseroth, Andres~Felipe Cruz-Salinas, Carlo Luschi, Samuel Weinbach, and Douglas Orr.
\newblock u-{$\mu$P}: The unit-scaled maximal update parametrization.
\newblock \emph{arXiv preprint arXiv:2407.17465}, 2024.

\bibitem[Cover \& Thomas(2006)Cover and Thomas]{coverthomas}
Thomas~M. Cover and Joy~A. Thomas.
\newblock \emph{Elements of Information Theory}.
\newblock Wiley, second edition, 2006.

\bibitem[Dao(2023)]{flashattention2}
Tri Dao.
\newblock {FlashAttention-2}: Faster attention with better parallelism and work partitioning.
\newblock \emph{arXiv preprint arXiv:2307.08691}, 2023.

\bibitem[Dao et~al.(2022)]{flashattention}
Tri Dao et~al.
\newblock {FlashAttention}: Fast and memory-efficient exact attention with {IO}-awareness.
\newblock \emph{arXiv preprint arXiv:2205.14135}, 2022.

\bibitem[{DeepSeek-AI}(2025)]{dsa}
{DeepSeek-AI}.
\newblock {DeepSeek-V3.2}: Pushing the frontier of open large language models.
\newblock \emph{arXiv preprint arXiv:2512.02556}, 2025.
\newblock {DSA} and its lightning indexer first described in the {DeepSeek-V3.2-Exp} technical report, September 2025.

\bibitem[{DeepSeek-AI}(2026)]{deepseekv4}
{DeepSeek-AI}.
\newblock {DeepSeek-V4}: Towards highly efficient million-token context intelligence.
\newblock \emph{arXiv preprint arXiv:2606.19348}, 2026.

\bibitem[Everett et~al.(2024)Everett, Xiao, Wortsman, Alemi, Novak, Liu, Gur, Sohl-Dickstein, Kaelbling, Lee, and Pennington]{everett2024scaling}
Katie Everett, Lechao Xiao, Mitchell Wortsman, Alexander~A. Alemi, Roman Novak, Peter~J. Liu, Izzeddin Gur, Jascha Sohl-Dickstein, Leslie~Pack Kaelbling, Jaehoon Lee, and Jeffrey Pennington.
\newblock Scaling exponents across parameterizations and optimizers.
\newblock In \emph{International Conference on Machine Learning}, 2024.

\bibitem[Fu et~al.(2026)Fu, Xiao, Dong, Han, and Villa]{sparda}
Yaosheng Fu, Guangxuan Xiao, Xin Dong, Song Han, and Oreste Villa.
\newblock {SparDA}: Sparse decoupled attention for efficient long-context {LLM} inference.
\newblock In \emph{Advances in Neural Information Processing Systems (NeurIPS)}, 2026.
\newblock arXiv:2606.04511.

\bibitem[Gao et~al.(2024)Gao, Zeng, Du, Cao, Zhou, Qi, Lai, So, Cao, Yang, and Yang]{seerattention}
Yizhao Gao, Zhichen Zeng, Dayou Du, Shijie Cao, Peiyuan Zhou, Jiaxing Qi, Junjie Lai, Hayden Kwok-Hay So, Ting Cao, Fan Yang, and Mao Yang.
\newblock {SeerAttention}: Learning intrinsic sparse attention in your {LLMs}.
\newblock \emph{arXiv preprint arXiv:2410.13276}, 2024.

\bibitem[Gao et~al.(2026)Gao, Guo, Cao, Xia, Cheng, Wang, Ma, Sun, Ye, Dong, So, Hua, Cao, Yang, and Yang]{seerattentionr}
Yizhao Gao, Shuming Guo, Shijie Cao, Yuqing Xia, Yu~Cheng, Lei Wang, Lingxiao Ma, Yutao Sun, Tianzhu Ye, Li~Dong, Hayden Kwok-Hay So, Yu~Hua, Ting Cao, Fan Yang, and Mao Yang.
\newblock {SeerAttention-R}: Sparse attention adaptation for long reasoning.
\newblock In \emph{International Conference on Learning Representations (ICLR)}, 2026.
\newblock arXiv:2506.08889.

\bibitem[{GLM-5 Team}(2026)]{glm5}
{GLM-5 Team}.
\newblock {GLM-5}: from vibe coding to agentic engineering.
\newblock \emph{arXiv preprint arXiv:2602.15763}, 2026.

\bibitem[{Google DeepMind}(2026)]{gemini38}
{Google DeepMind}.
\newblock Introducing {Gemini 3.8 Flash} and {3.8 Flash Cyber}.
\newblock \url{https://blog.google/innovation-and-ai/models-and-research/gemini-models/3-8-flash-and-3-8-flash-cyber/}, September 2026.

\bibitem[Grattafiori et~al.(2024)]{llama3}
Aaron Grattafiori et~al.
\newblock The {Llama 3} herd of models.
\newblock \emph{arXiv preprint arXiv:2407.21783}, 2024.

\bibitem[Hanna et~al.(2024)Hanna, Pezzelle, and Belinkov]{hanna2024faithfulness}
Michael Hanna, Sandro Pezzelle, and Yonatan Belinkov.
\newblock Have faith in faithfulness: Going beyond circuit overlap when finding model mechanisms.
\newblock In \emph{Conference on Language Modeling}, 2024.

\bibitem[Hanna et~al.(2025)Hanna, Belinkov, and Pezzelle]{hanna2025formal}
Michael Hanna, Yonatan Belinkov, and Sandro Pezzelle.
\newblock Are formal and functional linguistic mechanisms dissociated in language models?
\newblock \emph{Computational Linguistics}, 2025.
\newblock arXiv:2503.11302.

\bibitem[Hooper et~al.(2025)Hooper, Zhao, Manolache, Kim, Mahoney, Shao, Keutzer, and Gholami]{multipole}
Coleman Hooper, Sebastian Zhao, Luca Manolache, Sehoon Kim, Michael~W. Mahoney, Yakun~Sophia Shao, Kurt Keutzer, and Amir Gholami.
\newblock Multipole attention for efficient long context reasoning.
\newblock \emph{arXiv preprint arXiv:2506.13059}, 2025.

\bibitem[Hsieh et~al.(2024)Hsieh, Sun, Kriman, Acharya, Rekesh, Jia, Zhang, and Ginsburg]{ruler}
Cheng-Ping Hsieh, Simeng Sun, Samuel Kriman, Shantanu Acharya, Dima Rekesh, Fei Jia, Yang Zhang, and Boris Ginsburg.
\newblock {RULER}: What's the real context size of your long-context language models?
\newblock \emph{arXiv preprint arXiv:2404.06654}, 2024.

\bibitem[Huang et~al.(2025)Huang, Wang, Han, Zhao, Su, Sun, Lyu, Zhao, Wang, Xiao, Han, and Liu]{nosa}
Yuxiang Huang, Pengjie Wang, Jicheng Han, Weilin Zhao, Zhou Su, Ao~Sun, Hongya Lyu, Hengyu Zhao, Yudong Wang, Chaojun Xiao, Xu~Han, and Zhiyuan Liu.
\newblock {NOSA}: Native and offloadable sparse attention.
\newblock \emph{arXiv preprint arXiv:2510.13602}, 2025.

\bibitem[Huang et~al.(2026)Huang, Gon{\c{c}}alves, Alvetreti, Li, Han, Ponti, Martins, and Treviso]{dashattention}
Yuxiang Huang, Nuno M.~T. Gon{\c{c}}alves, Federico Alvetreti, Lei Li, Xu~Han, Edoardo~M. Ponti, Andr{\'e} F.~T. Martins, and Marcos~V. Treviso.
\newblock Dashattention: Differentiable and adaptive sparse hierarchical attention.
\newblock In \emph{Advances in Neural Information Processing Systems (NeurIPS)}, 2026.
\newblock arXiv:2605.18753.

\bibitem[Hwang(2026)]{locks}
Junsung Hwang.
\newblock Locks: Page-local compact key summaries for efficient long-context decoding.
\newblock \emph{arXiv preprint arXiv:2607.24555}, 2026.

\bibitem[Ji et~al.(2026)Ji, Zhang, Fu, and Cui]{sale}
Xiaodong Ji, Hailin Zhang, Fangcheng Fu, and Bin Cui.
\newblock Sale : Low-bit estimation for efficient sparse attention in long-context llm prefilling.
\newblock In \emph{International Conference on Machine Learning (ICML)}, 2026.
\newblock arXiv:2505.24179.

\bibitem[Jiang et~al.(2024)]{minference}
Huiqiang Jiang et~al.
\newblock {MInference} 1.0: Accelerating pre-filling for long-context {LLMs} via dynamic sparse attention.
\newblock \emph{arXiv preprint arXiv:2407.02490}, 2024.

\bibitem[Jie et~al.(2025)Jie, Tang, Han, Deng, and Han]{specache}
Shibo Jie, Yehui Tang, Kai Han, Zhi-Hong Deng, and Jing Han.
\newblock {SpeCache}: Speculative key-value caching for efficient generation of {LLMs}.
\newblock \emph{arXiv preprint arXiv:2503.16163}, 2025.

\bibitem[Kalyanarangan(2026)]{fathom}
Vivek Kalyanarangan.
\newblock Fathom: Per-query read depth for sparse decoding over offloaded kv caches.
\newblock \emph{arXiv preprint arXiv:2609.17652}, 2026.

\bibitem[Kate et~al.(2025)Kate, Pedapati, Basu, Rizk, Chenthamarakshan, Chaudhury, Agarwal, and Abdelaziz]{longfunceval}
Kiran Kate, Tejaswini Pedapati, Kinjal Basu, Yara Rizk, Vijil Chenthamarakshan, Subhajit Chaudhury, Mayank Agarwal, and Ibrahim Abdelaziz.
\newblock {LongFuncEval}: Measuring the effectiveness of long context models for function calling.
\newblock \emph{arXiv preprint arXiv:2505.10570}, 2025.

\bibitem[{Kimi Team}(2026)]{kimik3}
{Kimi Team}.
\newblock {Kimi K3}: Open frontier intelligence.
\newblock \emph{arXiv preprint arXiv:2607.24653}, 2026.

\bibitem[Kram{\'a}r et~al.(2024)Kram{\'a}r, Lieberum, Shah, and Nanda]{kramar2024atpstar}
J{\'a}nos Kram{\'a}r, Tom Lieberum, Rohin Shah, and Neel Nanda.
\newblock {AtP*}: An efficient and scalable method for localizing {LLM} behaviour to components.
\newblock \emph{arXiv preprint arXiv:2403.00745}, 2024.

\bibitem[Lai et~al.(2026)Lai, Xu, Yang, Chen, Xu, Zeng, Li, Sun, Zhu, Zhang, Hu, Li, et~al.]{msa}
Xunhao Lai, Weiqi Xu, Yufeng Yang, Qiaorui Chen, Yang Xu, Lunbin Zeng, Xiaolong Li, Haohai Sun, Haichao Zhu, Vito Zhang, Jinkai Hu, Jiayao Li, et~al.
\newblock Minimax sparse attention.
\newblock \emph{arXiv preprint arXiv:2606.13392}, 2026.

\bibitem[Li et~al.(2025)Li, Guo, Li, Zhang, Li, Li, Tao, Liu, Tao, Zhu, and Jin]{longcodeu}
Jia Li, Xuyuan Guo, Lei Li, Kechi Zhang, Ge~Li, Jia Li, Zhengwei Tao, Fang Liu, Chongyang Tao, Yuqi Zhu, and Zhi Jin.
\newblock {LongCodeU}: Benchmarking long-context language models on long code understanding.
\newblock \emph{arXiv preprint arXiv:2503.04359}, 2025.

\bibitem[Li et~al.(2024)]{snapkv}
Yuhong Li et~al.
\newblock {SnapKV}: {LLM} knows what you are looking for before generation.
\newblock \emph{arXiv preprint arXiv:2404.14469}, 2024.

\bibitem[Lin et~al.(2025)Lin, Tang, Yang, Wang, Tang, Tian, Stoica, Han, and Gao]{twilight}
Chaofan Lin, Jiaming Tang, Shuo Yang, Hanshuo Wang, Tian Tang, Boyu Tian, Ion Stoica, Song Han, and Mingyu Gao.
\newblock Twilight: Adaptive attention sparsity with hierarchical top-$p$ pruning.
\newblock In \emph{Advances in Neural Information Processing Systems (NeurIPS)}, 2025.
\newblock arXiv:2502.02770.

\bibitem[Lin et~al.(2026)Lin, Li, Xing, He, Cui, Li, Ding, Zhou, and Tang]{dynamicheads}
Yuping Lin, Zitao Li, Yue Xing, Pengfei He, Yingqian Cui, Yaliang Li, Bolin Ding, Jingren Zhou, and Jiliang Tang.
\newblock Retrieval heads are dynamic.
\newblock \emph{arXiv preprint arXiv:2602.11162}, 2026.

\bibitem[Liu et~al.(2026{\natexlab{a}})Liu, Wang, Chen, Gong, Yuan, Zhao, Feng, Chen, and Guo]{absparse}
Di~Liu, Ruitian Wang, Chen Chen, Mingliang Gong, Yongjie Yuan, Han Zhao, Yu~Feng, Quan Chen, and Minyi Guo.
\newblock {AB-Sparse}: Sparse attention with adaptive block size for accurate and efficient long-context inference.
\newblock \emph{arXiv preprint arXiv:2605.12110}, 2026{\natexlab{a}}.

\bibitem[Liu et~al.(2024{\natexlab{a}})Liu, Tian, Daita, Wei, Ding, Wang, Yang, and Zhang]{repoqa}
Jiawei Liu, Jia~Le Tian, Vijay Daita, Yuxiang Wei, Yifeng Ding, Yuhan~Katherine Wang, Jun Yang, and Lingming Zhang.
\newblock {RepoQA}: Evaluating long context code understanding.
\newblock \emph{arXiv preprint arXiv:2406.06025}, 2024{\natexlab{a}}.

\bibitem[Liu et~al.(2026{\natexlab{b}})Liu, Ning, Li, Liu, Song, Zhang, He, and Qiu]{ffd}
Zhigeng Liu, Zhiyuan Ning, Ruixiao Li, Xiaoran Liu, Yuerong Song, Min Zhang, Ziwei He, and Xipeng Qiu.
\newblock Faster than flash: Exploiting attention sparsity for efficient long-context decoding.
\newblock \emph{arXiv preprint arXiv:2609.00097}, 2026{\natexlab{b}}.

\bibitem[Liu et~al.(2024{\natexlab{b}})]{kivi}
Zirui Liu et~al.
\newblock {KIVI}: A tuning-free asymmetric 2bit quantization for {KV} cache.
\newblock \emph{arXiv preprint arXiv:2402.02750}, 2024{\natexlab{b}}.

\bibitem[Lu et~al.(2025)]{moba}
Enzhe Lu et~al.
\newblock {MoBA}: Mixture of block attention for long-context {LLMs}.
\newblock \emph{arXiv preprint arXiv:2502.13189}, 2025.

\bibitem[Marks et~al.(2025)Marks, Rager, Michaud, Belinkov, Bau, and Mueller]{marks2025sfc}
Samuel Marks, Can Rager, Eric~J. Michaud, Yonatan Belinkov, David Bau, and Aaron Mueller.
\newblock Sparse feature circuits: Discovering and editing interpretable causal graphs in language models.
\newblock In \emph{International Conference on Learning Representations}, 2025.

\bibitem[Max(1960)]{max1960}
Joel Max.
\newblock Quantizing for minimum distortion.
\newblock \emph{IRE Transactions on Information Theory}, 6\penalty0 (1):\penalty0 7--12, 1960.

\bibitem[{MiniCPM Team}(2025)]{minicpm4}
{MiniCPM Team}.
\newblock {MiniCPM4}: Ultra-efficient {LLMs} on end devices.
\newblock \emph{arXiv preprint arXiv:2506.07900}, 2025.

\bibitem[Nanda(2023)]{nanda2023atp}
Neel Nanda.
\newblock Attribution patching: Activation patching at industrial scale.
\newblock \url{https://www.neelnanda.io/mechanistic-interpretability/attribution-patching}, 2023.

\bibitem[{OpenAI}(2025)]{mrcr}
{OpenAI}.
\newblock {OpenAI MRCR}: Long context multiple needle in a haystack benchmark.
\newblock \url{https://huggingface.co/datasets/openai/mrcr}, 2025.
\newblock MIT licence; revision f4c69fa.

\bibitem[{OpenAI}(2026)]{gpt6astra}
{OpenAI}.
\newblock {GPT-6 Astra}: A new generation of intelligence.
\newblock \url{https://openai.com/index/gpt-6-astra/}, 2026.

\bibitem[Ren et~al.(2026)Ren, Sun, Yi, and Yang]{cfaudit}
Xingyu Ren, Youran Sun, Chugang Yi, and Haizhao Yang.
\newblock Understanding sparse attention selectivity in long-context foundation models via counterfactual evaluation.
\newblock \emph{arXiv preprint arXiv:2608.01676}, 2026.

\bibitem[Shah et~al.(2024)]{flashattention3}
Jay Shah et~al.
\newblock {FlashAttention-3}: Fast and accurate attention with asynchrony and low-precision.
\newblock \emph{arXiv preprint arXiv:2407.08608}, 2024.

\bibitem[Singhania et~al.(2024)Singhania, Singh, He, Feizi, and Bhatele]{loki}
Prajwal Singhania, Siddharth Singh, Shwai He, Soheil Feizi, and Abhinav Bhatele.
\newblock Loki: Low-rank keys for efficient sparse attention.
\newblock In \emph{Advances in Neural Information Processing Systems}, 2024.

\bibitem[Song et~al.(2026)Song, Wei, and Peng]{rbs}
Chuxu Song, Jiuqi Wei, and Zhencan Peng.
\newblock {RBS-Attention}: Radius-bounded sparse prefill for long-context large language models.
\newblock \emph{arXiv preprint arXiv:2609.20971}, 2026.

\bibitem[Sun et~al.(2025)Sun, Li, Zhang, Pan, Dong, Guo, and Wang]{effattnsurvey}
Yutao Sun, Zhenyu Li, Yike Zhang, Tengyu Pan, Bowen Dong, Yuyi Guo, and Jianyong Wang.
\newblock Efficient attention mechanisms for large language models: A survey.
\newblock \emph{arXiv preprint arXiv:2507.19595}, 2025.

\bibitem[Sundararajan et~al.(2017)Sundararajan, Taly, and Yan]{sundararajan2017ig}
Mukund Sundararajan, Ankur Taly, and Qiqi Yan.
\newblock Axiomatic attribution for deep networks.
\newblock In \emph{International Conference on Machine Learning}, 2017.

\bibitem[Syed et~al.(2023)Syed, Rager, and Conmy]{syed2023atp}
Aaquib Syed, Can Rager, and Arthur Conmy.
\newblock Attribution patching outperforms automated circuit discovery.
\newblock \emph{arXiv preprint arXiv:2310.10348}, 2023.

\bibitem[Tang et~al.(2024{\natexlab{a}})Tang, Lin, Lin, Han, Hong, Yao, and Wang]{razorattention}
Hanlin Tang, Yang Lin, Jing Lin, Qingsen Han, Shikuan Hong, Yiwu Yao, and Gongyi Wang.
\newblock {RazorAttention}: Efficient {KV} cache compression through retrieval heads.
\newblock \emph{arXiv preprint arXiv:2407.15891}, 2024{\natexlab{a}}.

\bibitem[Tang et~al.(2024{\natexlab{b}})Tang, Zhao, Zhu, Xiao, Kasikci, and Han]{quest}
Jiaming Tang, Yilong Zhao, Kan Zhu, Guangxuan Xiao, Baris Kasikci, and Song Han.
\newblock Quest: Query-aware sparsity for efficient long-context {LLM} inference.
\newblock \emph{arXiv preprint arXiv:2406.10774}, 2024{\natexlab{b}}.

\bibitem[Tang et~al.(2026)Tang, Qiu, Yang, Hong, Xiang, Liu, Dang, Li, and Zhang]{elasticattn}
Zecheng Tang, Quantong Qiu, Yi~Yang, Zhiyi Hong, Haiya Xiang, Kebin Liu, Qingqing Dang, Juntao Li, and Min Zhang.
\newblock Elastic attention: Test-time adaptive sparsity ratios for efficient transformers.
\newblock \emph{arXiv preprint arXiv:2601.17367}, 2026.

\bibitem[Tian et~al.(2026)Tian, Ghai, Neelam, Vasania, and Mishra]{cobs}
Alexander Tian, Aditya Ghai, Sanjit Neelam, Zaal Vasania, and Akshay Mishra.
\newblock {COBS}: Cumulant order block sparse attention.
\newblock \emph{arXiv preprint arXiv:2607.09052}, 2026.

\bibitem[Tillet et~al.(2019)Tillet, Kung, and Cox]{triton}
Philippe Tillet, Hsiang-Tsung Kung, and David Cox.
\newblock Triton: An intermediate language and compiler for tiled neural network computations.
\newblock In \emph{Proceedings of the 3rd ACM SIGPLAN International Workshop on Machine Learning and Programming Languages (MAPL)}, 2019.

\bibitem[Wang et~al.(2025)Wang, Liu, Wang, Ren, Deng, Hu, Chen, and Yang]{fier}
Dongwei Wang, Zijie Liu, Song Wang, Yuxin Ren, Jianing Deng, Jingtong Hu, Tianlong Chen, and Huanrui Yang.
\newblock {FIER}: Fine-grained and efficient {KV} cache retrieval for long-context {LLM} inference.
\newblock \emph{arXiv preprint arXiv:2508.08256}, 2025.

\bibitem[Wang et~al.(2026{\natexlab{a}})Wang, Razanajato, Zhang, Yuan, and Liu]{oracleprefill}
Hongxing Wang, Harenome Razanajato, Zhen Zhang, Yujie Yuan, and Hongsheng Liu.
\newblock How much dense attention is necessary? {Oracle-Guided} sparse prefill for full/{GQA} layers in hybrid long-context models.
\newblock \emph{arXiv preprint arXiv:2606.07703}, 2026{\natexlab{a}}.

\bibitem[Wang et~al.(2022)Wang, Variengien, Conmy, Shlegeris, and Steinhardt]{wang2022ioi}
Kevin Wang, Alexandre Variengien, Arthur Conmy, Buck Shlegeris, and Jacob Steinhardt.
\newblock Interpretability in the wild: a circuit for indirect object identification in {GPT-2} small.
\newblock \emph{arXiv preprint arXiv:2211.00593}, 2022.

\bibitem[Wang et~al.(2026{\natexlab{b}})Wang, Wang, Liu, Liu, Chu, Song, and Qiu]{prism}
Xinghao Wang, Pengyu Wang, Xiaoran Liu, Fangxu Liu, Jason Chu, Kai Song, and Xipeng Qiu.
\newblock Prism: Spectral-aware block-sparse attention.
\newblock In \emph{International Conference on Machine Learning (ICML)}, 2026{\natexlab{b}}.
\newblock arXiv:2602.08426.

\bibitem[Wu et~al.(2024{\natexlab{a}})Wu, Pan, Wang, Chen, Bai, Wang, Fu, Wang, and Xiong]{tokenselect}
Wei Wu, Zhuoshi Pan, Chao Wang, Liyi Chen, Yunchu Bai, Tianfu Wang, Kun Fu, Zheng Wang, and Hui Xiong.
\newblock Tokenselect: Efficient long-context inference and length extrapolation for llms via dynamic token-level kv cache selection.
\newblock \emph{arXiv preprint arXiv:2411.02886}, 2024{\natexlab{a}}.

\bibitem[Wu et~al.(2024{\natexlab{b}})]{retrievalhead}
Wenhao Wu et~al.
\newblock Retrieval head mechanistically explains long-context factuality.
\newblock \emph{arXiv preprint arXiv:2404.15574}, 2024{\natexlab{b}}.

\bibitem[Xiao et~al.(2023)Xiao, Tian, Chen, Han, and Lewis]{streamingllm}
Guangxuan Xiao, Yuandong Tian, Beidi Chen, Song Han, and Mike Lewis.
\newblock Efficient streaming language models with attention sinks.
\newblock \emph{arXiv preprint arXiv:2309.17453}, 2023.

\bibitem[Xiao et~al.(2025)Xiao, Guo, Mazaheri, and Han]{optmoba}
Guangxuan Xiao, Junxian Guo, Kasra Mazaheri, and Song Han.
\newblock Optimizing mixture of block attention.
\newblock \emph{arXiv preprint arXiv:2511.11571}, 2025.

\bibitem[Xiao et~al.(2024)]{duoattention}
Guangxuan Xiao et~al.
\newblock {DuoAttention}: Efficient long-context {LLM} inference with retrieval and streaming heads.
\newblock \emph{arXiv preprint arXiv:2410.10819}, 2024.

\bibitem[Xu et~al.(2026)Xu, Meng, Jiang, Wang, Zhou, Wang, Wu, Pan, Tang, Pei, Liu, Yin, et~al.]{hisa}
Yufei Xu, Fanxu Meng, Fan Jiang, Yuxuan Wang, Ruijie Zhou, Zhaohui Wang, Jiexi Wu, Zhixin Pan, Xiaojuan Tang, Wenjie Pei, Tongxuan Liu, Di~Yin, et~al.
\newblock Hisa: Efficient hierarchical indexing for fine-grained sparse attention.
\newblock \emph{arXiv preprint arXiv:2603.28458}, 2026.

\bibitem[Yang et~al.(2025)Yang, Li, Yang, et~al.]{qwen3}
An~Yang, Anfeng Li, Baosong Yang, et~al.
\newblock {Qwen3} technical report.
\newblock \emph{arXiv preprint arXiv:2505.09388}, 2025.

\bibitem[Yang et~al.(2024)Yang, Sheng, Gonzalez, Stoica, and Zheng]{doublesparsity}
Shuo Yang, Ying Sheng, Joseph~E. Gonzalez, Ion Stoica, and Lianmin Zheng.
\newblock Post-training sparse attention with double sparsity.
\newblock \emph{arXiv preprint arXiv:2408.07092}, 2024.

\bibitem[Yang et~al.(2026)Yang, Zhang, {Zhangke}, Chen, Chen, Sun, Xu, Jin, and Tang]{selfindexing}
Xu~Yang, Jiapeng Zhang, {Zhangke}, Changjian Chen, Yuxin Chen, Feiqiang Sun, Chengguang Xu, Feng Jin, and Zhuo Tang.
\newblock Self-indexing attention for compression-compatible sparse long-context llm inference.
\newblock \emph{arXiv preprint arXiv:2609.13205}, 2026.

\bibitem[Yen et~al.(2024)]{helmet}
Howard Yen et~al.
\newblock {HELMET}: How to evaluate long-context language models effectively and thoroughly.
\newblock \emph{arXiv preprint arXiv:2410.02694}, 2024.

\bibitem[Yuan et~al.(2025)Yuan, Gao, Dai, Luo, Zhao, Zhang, Xie, Wei, Wang, Xiao, Wang, Ruan, Zhang, Liang, and Zeng]{nsa}
Jingyang Yuan, Huazuo Gao, Damai Dai, Junyu Luo, Liang Zhao, Zhengyan Zhang, Zhenda Xie, Y.~X. Wei, Lean Wang, Zhiping Xiao, Yuqing Wang, Chong Ruan, Ming Zhang, Wenfeng Liang, and Wangding Zeng.
\newblock Native sparse attention: Hardware-aligned and natively trainable sparse attention.
\newblock \emph{arXiv preprint arXiv:2502.11089}, 2025.

\bibitem[Zan et~al.(2026)Zan, Zhang, Tan, Liu, Wang, Li, et~al.]{longcatlsa}
Wen Zan, Jiaqi Zhang, Jianchao Tan, Hong Liu, Cunguang Wang, Xiang Li, et~al.
\newblock Longcat sparse attention: Taming the lightning via streaming-aware hierarchical cross-layer indexing.
\newblock \emph{arXiv preprint arXiv:2608.01662}, 2026.

\bibitem[Zhao et~al.(2026)Zhao, Zhou, Su, Xiao, Li, Li, Zhang, Zhao, Li, Huang, Sun, Han, et~al.]{infllmv2}
Weilin Zhao, Zihan Zhou, Zhou Su, Chaojun Xiao, Yuxuan Li, Yanghao Li, Yudi Zhang, Weilun Zhao, Zhen Li, Yuxiang Huang, Ao~Sun, Xu~Han, et~al.
\newblock {InfLLM-V2}: Dense-sparse switchable attention for seamless short-to-long adaptation.
\newblock In \emph{International Conference on Learning Representations (ICLR)}, 2026.
\newblock arXiv:2509.24663.

\bibitem[{Zhipu AI}(2026)]{glm53}
{Zhipu AI}.
\newblock {GLM-5.3}: Frontier coding with emergent cyber capabilities.
\newblock \url{https://z.ai/blog/glm-5.3}, 2026.

\bibitem[Zhou et~al.(2026)Zhou, Li, Tang, Li, Liu, Lan, Qu, Yao, and Ma]{rtpurbo}
Yanke Zhou, Yiduo Li, Hanlin Tang, Maohua Li, Kan Liu, Tao Lan, Lin Qu, Yuan Yao, and Xiaoxing Ma.
\newblock Full attention strikes back: Transferring full attention into sparse within hundred training steps.
\newblock \emph{arXiv preprint arXiv:2605.16928}, 2026.

\end{thebibliography}
\bibliographystyle{preprint}

\appendix
\setlength{\intextsep}{10pt plus 2pt minus 2pt}%
\raggedbottom%
\section{Setup}
\label{app:setup}
\label{app:harness}
\textbf{Checkpoints.} Table~\ref{tab:arch} gives the selector settings of the four \infllm{} checkpoints. Each
checkpoint's full attention setting loads the same weights without the release's sparse configuration.

\begin{table}[H]
\caption{\textbf{Selector settings of the lineage.} The four checkpoints share the MiniCPM4-8B backbone, 32 layers of 32
query heads and 2 key-value heads of dimension 128. The selector scores keys mean-pooled over windows of 32 at a stride
of 16, maps the window scores onto blocks of 64 tokens and always reads the initial block, a local window and the query's
own block. Tokens below the last column's position attend densely.}
\label{tab:arch}
\centering\vskip 6pt
\footnotesize
\setlength{\tabcolsep}{4pt}
\begin{tabular}{lccccc}
\toprule
Checkpoint & Local Window & Forced Blocks & Free Slots & Blocks per Token & Dense Below \\
\midrule
\infllm{}-8B & 2,048 & 34 & 62 & 96 & 8,192 \\
MiniCPM4-8B Instruct & 2,048 & 34 & 62 & 96 & 8,192 \\
\mcpm{} & 2,048 & 34 & 62 & 96 & 8,192 \\
InfLLMv2-8B Retraining & 1,024 & 18 & 46 & 64 & None \\
\bottomrule
 
\end{tabular}
\end{table}

\textbf{Protocol.} We use RULER \citep{ruler} at a pinned generator commit. An \mk{3} prompt is a haystack of lines that
pair a UUID key with a UUID value, one of which holds the queried key, and the last 128 prompt tokens hold the question.
Every cell reports string-match accuracy on 200 prompts of one generator seed, and every comparison is paired on the
same prompts. The first seed guided the design choices, five further seeds carry the main results and calibration
prompts come from other seeds.

\textbf{Fidelity and controls.} Every arm acts through a hook on the selector's output. A keep-only arm that fills the
free slots from the kernel's own score reproduces the released selector exactly, and every run checks its hook calls
against their expected counts. Negative controls run through the same path: the scorer in the other KV head of
the flagged layers, random free blocks in the flagged heads, mean-pooled keys under the scorer's criterion in the flagged
heads, the scorer in the 61 unflagged heads and full attention in three random unflagged heads drawn three times.

\section{The calibration probe}
\label{app:calib}
\label{app:fixgain}
\textbf{The forfeited attention.} The probe runs the shipped model with its own selector and teacher-forces the answer tokens after
the prompt's sparse prefill. It reads every quantity below from the queries and cached keys of that run. At decode step
$t$, the model must produce the needle token at context position $a$. The needle-containing block has index
$n=\lfloor a/L\rfloor$, for block length $L$. For a key-value head $h$, let $\mathcal P_h$ collect these pairs $(t,n)$
over all calibration prompts whose needle block is not forced. Let $\mathcal S^h_t$ be the head's shipped selection at
token $t$, and let $\mathcal E^h_t=\operatorname*{TopK}_{b\notin\mathcal F}(M^h_{t,b},\,K)$ be the selection that exact
mass makes at the same budget, with $M^h_{t,b}$ being the head's exact mass from \Eqref{eq:exact}. The probe records
the needle recall of a selection $\mathcal A$ and the head's mean needle mass,
\begin{equation}
r_h(\mathcal A)=\frac{1}{|\mathcal P_h|}\sum_{(t,n)\in\mathcal P_h}\mathbf{1}\bigl[n\in\mathcal A^h_t\bigr],\qquad
m_h=\frac{1}{|\mathcal P_h|}\sum_{(t,n)\in\mathcal P_h}\frac{M^h_{t,n}}{G},
\label{eq:probe}
\end{equation}
and the forfeited attention of a head is $F_h=\bigl(r_h(\mathcal E)-r_h(\mathcal S)\bigr)\,m_h$.

\textbf{The fix gain.} The probe runs the gate of \Eqref{eq:gate} at every position where the selector acts and
differentiates the log-probability of the needle tokens with respect to every gate at once. SeerAttention-R selects only
at decode, so its probe decodes the needle tokens one at a time with their keys and values kept in the graph,
recomputes each head's decode attention in PyTorch over the blocks the gate reads, and checks it against the kernel's
output, within 4.2 percent of its largest value. The other selectors act at every prompt token, so their probe runs one
teacher-forced pass over the prompt and the needle tokens through the releases' own kernels, which have backward passes,
with gradient checkpointing per layer. MoBA's kernel needed a wrapper for the backward pass of FlashAttention 2.7, checked
against MoBA's reference implementation, and NSA's release computes its compressed branch without gradients, a path the
probe adds back. Each derivative treats the selections as constants and recomputes them at every point of the path.
Summed over units, the integrals recover most of the log-probability gap over the calibration prompts
(Table~\ref{tab:calcost}). The rest is lost where selections switch along the path, and the midpoint rule on sixteen
points resolves the sharp transitions that the path shows on many prompts. The absolute value per prompt matters. With the first-order version, the signed
mean ranks L9.h0 of \infllm{}-8B last of the 64 key-value heads while its mean absolute fix gain is the second largest,
because fixing that head moves the answer strongly in both directions and the signs cancel.

\textbf{Choosing the score.} Three versions of the fix gain were compared before the choice, by a criterion fixed in
advance as the smallest geometric mean over the eight checkpoints of Table~\ref{tab:rule} of the count the rule keeps
(Table~\ref{tab:probevariants}). The first-order version takes the derivative at $\bm\alpha=\bm 0$ alone, holds the keys
and values of the cache constant and takes the absolute value per query head and token. The complete version keeps every
path and takes the absolute value per prompt, and the integrated version integrates the complete derivative along the
path. The integrated version keeps the fewest heads on average, though not on every checkpoint, and the other two remain
as ablations.

\textbf{The calibration set.} The haystack decides which heads a source can reveal, and scores of dense models rank
other heads (Table~\ref{tab:calibset}). A pooled score misranks a needle only against look-alike competitors
(Proposition~\ref{prop:pool}), so a calibration set must contain them. A scan that needs no probe singles out the same
three heads (Table~\ref{tab:scan}).

\textbf{How many heads.} Table~\ref{tab:lineage} runs the rule on the lineage. Leaving out the head of layer 16 on
\infllm{}-8B costs the scorer 2.8 points on average over the five evaluation seeds, and adding the fourth head by the fix
gain on \mcpm{}, the first head of layer 23, moves the scorer by less than half a point, so the lineage keeps the three
blind heads.

\begin{table}[H]
\caption{\textbf{Three versions of the fix gain.} The count the rule keeps, the first step of the $\sqrt2$ sweep that
closes 90 percent of the gap on calibration prompts, for the first-order, complete and integrated versions of the fix
gain, the last of them \Eqref{eq:fixgain}.}
\label{tab:probevariants}
\centering\vskip 6pt
\footnotesize
\begin{tabular}{lcccc}
\toprule
Model & Units & First Order & Complete & Integrated \\
\midrule
SeerAttention-R, Qwen3-8B & 288 & 10 & 18 & 18 \\
SeerAttention-R, Qwen3-4B & 288 & 11 & 15 & 9 \\
SeerAttention-R, Qwen3-14B & 320 & 5 & 5 & 5 \\
SeerAttention-R, R1-Distill-Qwen-14B & 384 & 49 & 16 & 22 \\
Dash-8B & 64 & 19 & 23 & 13 \\
NSA-8B & 64 & 5 & 4 & 4 \\
MoBA, Qwen3-8B & 1152 & 587 & 585 & 513 \\
MoBA, Qwen3-4B & 1152 & 388 & 604 & 433 \\
\midrule
Geometric mean & & 29.6 & 30.3 & 26.0 \\
\bottomrule
 
\end{tabular}
\end{table}

\begin{table}[H]
\caption{\textbf{What the calibration source and the score decide} on \infllm{}-8B. The heads that the forfeited attention
flags from each source and the three best heads by each score, with \mk{3} at 32K for the scorer at every token of those
heads. When no head is flagged, the scorer is the shipped selector.}
\label{tab:calibset}
\centering\vskip 6pt
\footnotesize
\begin{tabular}{llc}
\toprule
 & Heads & \mk{3} \\
\midrule
\multicolumn{3}{l}{\textit{Calibration source}} \\
\quad \mk{3}, the default & L9.h0, L16.h0, L22.h0 & 98.0 \\
\quad HELMET's JSON key-value data & L9.h0, L16.h0, L22.h0 & 98.0 \\
\quad \mk{2} & L9.h0, L22.h0 & 95.0 \\
\quad \mk{1}, an essay haystack & L3.h0, L6.h0, L9.h0 & 82.0 \\
\quad One needle among repeated filler sentences & None & 65.5 \\
\midrule
\multicolumn{3}{l}{\textit{Score that ranks the heads, on the default source}} \\
\quad Forfeited attention & L9.h0, L16.h0, L22.h0 & 98.0 \\
\quad Recall gap alone & L6.h0, L6.h1, L9.h0 & 82.0 \\
\quad Needle mass alone & L22.h0, L26.h0, L26.h1 & 80.5 \\
\quad Retrieval score \citep{retrievalhead} & L22.h0, L26.h0, L29.h1 & 81.0 \\
\bottomrule
 
\end{tabular}
\end{table}

\begin{table}[H]
\caption{\textbf{A scan without the probe.} The gain in \mk{3} at 32K when one key-value head of \infllm{}-8B gets full
attention and the other 63 keep the shipped selector.}
\label{tab:scan}
\centering\vskip 6pt
\footnotesize
\begin{tabular}{lc}
\toprule
Head & Gain over Sparse \\
\midrule
L9.h0 & $+15.5$ \\
L16.h0 & $+23.0$ \\
L22.h0 & $+17.0$ \\
Largest of the other 61 heads & $+2.0$ \\
\bottomrule
 
\end{tabular}
\end{table}

\begin{table}[H]
\caption{\textbf{The rule on the lineage.} The heads that the integrated fix gain ranks first on sixty calibration prompts,
the count at the first step of its sweep that closes 90 percent of the gap on the 200 calibration prompts and the share of
the gap closed there.}
\label{tab:lineage}
\centering\vskip 6pt
\footnotesize
\begin{tabular}{llcc}
\toprule
Checkpoint & First Four Heads by the Fix Gain & Heads Kept & Share of the Gap \\
\midrule
\infllm{}-8B & L22.h0, L9.h0, L16.h0, L29.h1 & 2 & 0.94 \\
MiniCPM4-8B Instruct & L16.h0, L9.h0, L22.h0, L29.h1 & 4 & 0.97 \\
\mcpm{} & L22.h0, L16.h0, L9.h0, L23.h0 & 3 & 1.00 \\
InfLLMv2-8B Retraining & L9.h0, L16.h0, L22.h0, L8.h1 & 3 & 0.97 \\
\bottomrule
 
\end{tabular}
\end{table}

\section{The resolution account}
\label{sec:theory}
A selector ranks blocks from a compact summary of the keys, which can give up resolution as pooled keys do, channels as
low-rank projections do or precision as quantized keys do. We use the notation of Section~\ref{sec:prelim}. A head's
log mass of a block is $f_b-\log Z$ with the block log-sum-exp $f_b=\log\sum_{j\in \mathcal{B}_b}e^{s_j}$, so it ranks
blocks by $f_b$. Exact mass selection keeps the $K$ free blocks of largest group mass $M_b$ in \Eqref{eq:exact}.

\begin{proposition}[Pooling has a blind window]
\label{prop:pool}
Let block $b$ have logits $s_0$ on all keys except one, the needle, with logit $s_0+\Delta$, and let
block $c$ have logit $s_0+\delta$ on every key, with $\Delta,\delta>0$. Exact mass ranks $b$ above
$c$ if and only if $\Delta>\log\bigl(1+L(e^{\delta}-1)\bigr)$, which holds whenever
$\Delta>\delta+\log L$. Any score that is an increasing function of the mean logit over a window
of $w$ keys ranks $c$ above $b$ whenever $\Delta<w\delta$. Every needle with
$\delta+\log L < \Delta < w\,\delta$
is therefore ranked correctly by exact mass and incorrectly by the pooled score.
\end{proposition}
\label{app:proofs}\emph{Proof.} \emph{(i)} By construction, $f_b=s_0+\log(L-1+e^{\Delta})$ and
$f_c=s_0+\delta+\log L$, so $f_b>f_c$ if and only if $e^{\Delta}>1+L(e^{\delta}-1)$. Since
$Le^{\delta}-\bigl(1+L(e^{\delta}-1)\bigr)=L-1\ge0$, $\Delta>\delta+\log L$ suffices. \emph{(ii)} A score is linear in
the key, so for the mean $\bar{\mathbf{k}}$ of a window $W$, $\mathbf{q}^\top\bar{\mathbf{k}}/\sqrt d=\frac1w\sum_{j\in W}s_j$
exactly. Pooling keys is averaging logits, which no nonlinearity downstream can undo. A window of $b$ that holds the
needle scores $s_0+\Delta/w$ and every window of $c$ scores $s_0+\delta$. \emph{(iii)} Any block score increasing in its
window scores, which includes the sum, the maximum and \infllm{}'s softmax over windows, ranks $c$ above $b$ whenever
$\Delta/w<\delta$. With overlapping windows, every window holding the needle still scores $s_0+\Delta/w$. $\square$

\textbf{The group.} \infllm{} scores a window by summing the scores of the query heads that share a key-value head, and
exact mass sums their masses. When the other heads of the group are flat over both blocks, both sums rank $b$ and $c$ as
the head that reads the needle does, so the proposition holds for the group. \infllm{}'s pooled keys are plain means of
the same rotated keys that attention reads, with no learned projection, scored by the same rotated queries, so step (ii)
holds for it exactly. Its first stage is a softmax over windows of these pooled keys, and its later stages are monotone
in the window scores. The proposition therefore applies to it.

\textbf{The window on the data.} What separates the heads is the rest of the needle's block. In the heads whose selector
finds the needle, its other keys sit well above the median. In the blind heads of layers 9 and 16, they sit at or below
it, the proposition's case of one raised key in a flat block (Table~\ref{tab:anatomy}).

\begin{proposition}[Quantization moves a block score by at most the largest logit error]
\label{prop:quant}
Let $\hat{\mathbf{k}}_j=\mathbf{k}_j+\boldsymbol{\varepsilon}_j$ be the keys used for scoring, with logit errors
$e_j=\mathbf{q}^\top\boldsymbol{\varepsilon}_j/\sqrt d$. Then, $|\hat f_b-f_b|\le\max_{j\in \mathcal{B}_b}|e_j|$ for every block, and
the order of two blocks is preserved whenever their margin exceeds $2\max_j|e_j|$. If the errors are
uncorrelated across channels with $\mathbb{E}\,\varepsilon_{jc}^2=D\,\sigma_c^2$, where $\sigma_c^2$
is the variance of channel $c$ over the keys, then
$\mathbb{E}\,e_j^2=D\,\sigma_s^2$ with $\sigma_s^2=\tfrac1d\sum_c q_c^2\sigma_c^2$, which is the
variance of the exact logits over keys when the key channels are uncorrelated.
\end{proposition}
\emph{Proof.} The gradient of $\log\sum_j e^{s_j}$ is the softmax of $s$, a probability
vector, so the map is 1-Lipschitz from the supremum norm and $|\hat f_b-f_b|\le\max_{j\in \mathcal{B}_b}|e_j|$. Two
block scores that differ by more than $2\max_j|e_j|$ each move by less than half the gap. With
$e_j=\sum_c q_c\varepsilon_{jc}/\sqrt d$ and uncorrelated channel errors,
$\mathbb{E}\,e_j^2=\frac1d\sum_c q_c^2D\sigma_c^2=D\sigma_s^2$, the variance of the exact logits over keys times $D$
when the channels are uncorrelated, independent of the query, the head and the length. $\square$

For the per-channel uniform quantizer of \Eqref{eq:quant}, $D$ is the distortion of the quantizer on the standardized
keys, which the Gaussian step fixes at each bit width \citep{max1960}. Each additional bit roughly halves the noise, and
the measured noise of the five probed heads matches the prediction at every bit width (Table~\ref{tab:noise}). At 64K, the
knee moves to three or four bits, which is why the scorer uses four. Averaging keys instead lowers accuracy with every
doubling of the window.

\begin{table}[H]
\caption{\textbf{Proposition~\ref{prop:pool} on the probed tokens} of \infllm{}-8B at 32K. ``Sparse Recall'' is the
share of probed tokens on which the shipped selector keeps the needle block. In each head's query head with the most
needle mass, ``Needle Key'' is the median margin of the needle key over the other keys of its block and ``Rest of Block''
the median of those other keys over the head's median window, in nats, and ``Exact Keeps'' and ``Pooled Loses'' are the
shares of tokens on which exact block mass keeps the needle block and the pooled score loses it. The last two columns
give how often the three query heads with the most needle mass rank the needle block among their own top free blocks.}
\label{tab:anatomy}
\label{tab:probed}
\centering\vskip 6pt
\footnotesize
\setlength{\tabcolsep}{3.5pt}
\fitwidth{%
\begin{tabular}{llccccccc}
\toprule
 & & & \multicolumn{2}{c}{Nats} & \multicolumn{2}{c}{Block Score} & \multicolumn{2}{c}{Top Query Heads} \\
\cmidrule(lr){4-5}\cmidrule(lr){6-7}\cmidrule(lr){8-9}
Head & Flagged & Sparse Recall & Needle Key & Rest of Block & Exact Keeps & Pooled Loses & Token Level & Pooled \\
\midrule
L9.h0 & Yes & 0.19 & 21.4 & $-0.6$ & 0.89 & 0.65 & 0.74 & 0.18 \\
L16.h0 & Yes & 0.24 & 21.6 & $-0.1$ & 0.86 & 0.65 & 0.71 & 0.25 \\
L22.h0 & Yes & 0.73 & 28.2 & 3.8 & 1.00 & 0.25 & 0.97 & 0.75 \\
L23.h1 & No & 0.83 & 7.4 & 4.4 & 0.94 & 0.07 & 0.85 & 0.85 \\
L26.h0 & No & 0.98 & 18.4 & 11.7 & 1.00 & 0.02 & 1.00 & 0.99 \\
\bottomrule
 
\end{tabular}}
\end{table}

\begin{figure}[H]
\centering
\includegraphics[width=\linewidth]{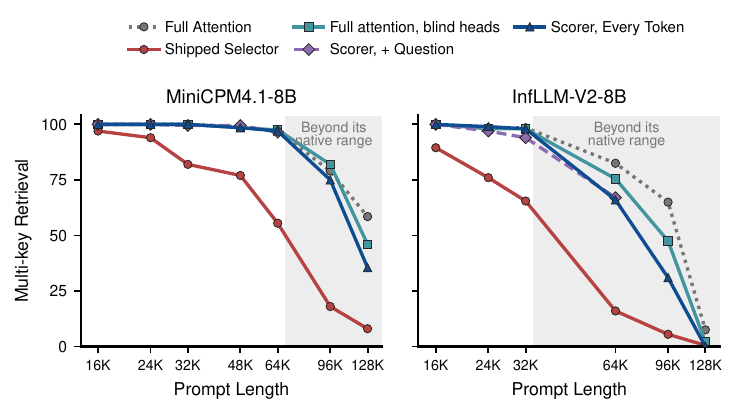}
\caption{\textbf{The deficit grows with the length} (\mk{3} at the shipped budget, 200 prompts per point, 100 at 96K and 128K). The 16K, 24K
and 48K points are draws of the same generator seed as the first seed at 32K and 64K, and the flagged heads are those
calibrated at 32K. The shaded ranges lie beyond each model's native range.}
\label{fig:lensweep}
\end{figure}

\begin{table}[H]
\caption{\textbf{Precision against resolution} on \infllm{}-8B at 32K. Left, Proposition~\ref{prop:quant} against the
probe on the five heads of Table~\ref{tab:anatomy}. Predicted noise is $\sqrt D$ of the optimal uniform quantizer of a unit
Gaussian, measured noise is the standard deviation of the logit error over that of the exact logits, and recall is the
needle recall of the per-head criterion on the quantized keys minus that of exact mass. Right, keys averaged over windows
of $w$ tokens. \mk{3} is the criterion at the decode tokens of the flagged heads on each kind of key.}
\label{tab:noise}
\label{tab:pooling}
\centering\vskip 6pt
\footnotesize
\setlength{\tabcolsep}{4pt}
\begin{tabular}[t]{ccccc}
\toprule
Bits & Predicted Noise & Measured Noise & Recall & \mk{3} \\
\midrule
1 & 0.60 & 0.60 & $-0.12$ & 81.0 \\
2 & 0.34 & 0.35 & $-0.02$ & 90.5 \\
3 & 0.19 & 0.20 & $-0.01$ & 90.0 \\
4 & 0.11 & 0.11 & $+0.00$ & 90.0 \\
\bottomrule
 
\end{tabular}
\hspace{2em}
\begin{tabular}[t]{cc}
\toprule
$w$ & \mk{3} \\
\midrule
1 & 89.5 \\
4 & 86.0 \\
8 & 84.0 \\
16 & 71.5 \\
64 & 59.5 \\
\bottomrule
 
\end{tabular}
\end{table}

\section{The threshold rule}
\label{app:rule}

\begin{table}[H]
\caption{\textbf{Share and step.} The count the integrated rule keeps at each share with the step $\sqrt2$ and at each
step with the share 0.9. MoBA's sweep is not monotone in the count.}
\label{tab:ruleablation}
\centering\vskip 6pt
\footnotesize
\begin{tabular}{lcccccc}
\toprule
 & \multicolumn{4}{c}{Share, Step $\sqrt2$} & \multicolumn{2}{c}{Step, Share 0.9} \\
\cmidrule(lr){2-5}\cmidrule(lr){6-7}
Model & 0.80 & 0.85 & 0.90 & 0.95 & $2$ & $2^{1/4}$ \\
\midrule
SeerAttention-R, Qwen3-8B & 11 & 11 & 18 & 18 & 29 & 14 \\
SeerAttention-R, Qwen3-4B & 9 & 9 & 9 & 22 & 15 & 8 \\
SeerAttention-R, Qwen3-14B & 5 & 5 & 5 & 5 & 7 & 5 \\
SeerAttention-R, R1-Distill-Qwen-14B & 12 & 22 & 22 & 33 & 22 & 22 \\
Dash-8B & 13 & 13 & 13 & 22 & 22 & 11 \\
NSA-8B & 4 & 4 & 4 & 4 & 4 & 4 \\
MoBA, Qwen3-8B & 323 & 513 & 513 & 513 & 513 & 460 \\
MoBA, Qwen3-4B & 182 & 182 & 433 & 816 & 433 & 246 \\
\bottomrule
 
\end{tabular}
\end{table}

\begin{table}[H]
\caption{\textbf{The calibration per checkpoint} on one H100, over its 200 calibration prompts. The sweep step is the
median over the steps that were run. ``Gap Recovered'' is the share of the log-probability gap between the shipped selector
and full attention that the integrated gains of all units sum to.}
\label{tab:calcost}
\centering\vskip 6pt
\footnotesize
\begin{tabular}{lcccc}
\toprule
 & \multicolumn{2}{c}{Probe (GPU Hours)} & Sweep Step & Gap \\
\cmidrule(lr){2-3}
Model & Integrated & First Order & (GPU Minutes) & Recovered \\
\midrule
SeerAttention-R, Qwen3-8B & 1.6 & 0.3 & 11 & 0.79 \\
SeerAttention-R, Qwen3-4B & 1.6 & 0.2 & 10 & 0.87 \\
SeerAttention-R, Qwen3-14B & 2.0 & 0.3 & 16 & 0.89 \\
SeerAttention-R, R1-Distill-Qwen-14B & 2.2 & 0.4 & 17 & 0.76 \\
Dash-8B & 6.0 & 0.4 & 14 & 0.61 \\
NSA-8B & 7.8 & 0.5 & 11 & 0.82 \\
MoBA, Qwen3-8B & 12.4 & 0.8 & 12 & 0.77 \\
MoBA, Qwen3-4B & 11.1 & 0.7 & 39 & 0.84 \\
\bottomrule
 
\end{tabular}
\end{table}

\section{Further results}
\label{app:ablations}
\label{app:cost}
\label{app:gating}
\label{app:real}

\begin{table}[H]
\caption{\textbf{Where fixing the selector stops.} \mk{3} on the first seed. Exact mass selects at every token and keeps
the shipped budget. Where it falls short of full attention, the budget binds.}
\label{tab:bounds}
\centering\vskip 6pt
\footnotesize
\setlength{\tabcolsep}{4pt}
\begin{tabular}{lccccc}
\toprule
 & & \multicolumn{2}{c}{Exact Mass} & Full Attention, & \\
\cmidrule(lr){3-4}
Model & Sparse & Flagged Heads & Every Head & Flagged Heads & Dense \\
\midrule
\nosa{}-8B, 32K & 18.5 & 90.0 & 94.0 & 94.0 & 94.0 \\
InfLLMv2-8B Retraining, 32K & 23.5 & 82.5 & 85.0 & 90.5 & 92.0 \\
\infllm{}-8B, 64K & 16.0 & 64.0 & 69.0 & 75.5 & 82.5 \\
\bottomrule
 
\end{tabular}
\end{table}

\begin{table}[H]
\caption{\textbf{Aggregation tasks on \mcpm{} at 64K.} RULER's variable tracking and common word extraction on the first
seed. Each fix acts in the three flagged heads or in their layers.}
\label{tab:aggregation}
\centering\vskip 6pt
\footnotesize
\setlength{\tabcolsep}{4pt}
\begin{tabular}{lccccc}
\toprule
Task & Dense & Sparse & \begin{tabular}[b]{@{}c@{}}Full Attention,\\Flagged Heads\end{tabular} & \begin{tabular}[b]{@{}c@{}}4x Budget,\\Flagged Layers\end{tabular} & \begin{tabular}[b]{@{}c@{}}Scorer,\\Every Token\end{tabular} \\
\midrule
Variable tracking & 46.9 & \textbf{74.4} & 60.9 & 64.1 & \underline{74.2} \\
Common word extraction & 40.5 & 11.8 & 11.6 & \underline{13.5} & \textbf{13.8} \\
\bottomrule
 
\end{tabular}
\end{table}

\begin{table}[H]
\caption{\textbf{Real data.} LongFuncEval \citep{longfunceval}, LongCodeU \citep{longcodeu}, RepoQA \citep{repoqa},
OpenAI MRCR \citep{mrcr} and OOLONG \citep{oolong}, each with its own data, prompts and metric. The scorer acts at every
token of the three flagged heads.}
\label{tab:real}
\centering\vskip 6pt
\footnotesize
\begin{tabular}{llccc}
\toprule
Benchmark & Model & Dense & Sparse & \begin{tabular}[b]{@{}c@{}}Scorer,\\Every Token\end{tabular} \\
\midrule
LongFuncEval, 32K & \infllm{}-8B & 62.9 & 58.4 & 60.0 \\
LongFuncEval, 64K &  & 48.3 & 39.7 & 46.1 \\
\midrule
LongCodeU & \mcpm{} & 75.1 & 74.1 & 74.3 \\
RepoQA, 16K & & 84.0 & 82.0 & 82.6 \\
RepoQA, 32K & & 79.0 & 78.4 & 79.4 \\
OpenAI MRCR, 16K to 32K & & 19.6 & 19.9 & 19.3 \\
OOLONG, 16K and 32K & & 26.7 & 24.9 & 26.0 \\
\bottomrule
 
\end{tabular}
\end{table}

\begin{table}[H]
\caption{\textbf{Latency beside Figure~\ref{fig:latency}}, \infllm{}-8B on one H100. Left, the time per output token at
batch size one in CPM.cu, to which the scorer is not ported. Right, the attention time of one prefill over all 32 layers
in the release's kernels, with the scorer at every prompt token.}
\label{tab:latency}
\centering\vskip 6pt
\footnotesize
\setlength{\tabcolsep}{4pt}
\begin{tabular}{lccccccc}
\toprule
 & \multicolumn{3}{c}{ms per Output Token} & \multicolumn{4}{c}{Prefill Attention (s)} \\
\cmidrule(lr){2-4}\cmidrule(lr){5-8}
Context & Dense & Sparse & \begin{tabular}[b]{@{}c@{}}Full Attention,\\Flagged Heads\end{tabular} & Dense & Sparse & \begin{tabular}[b]{@{}c@{}}Full Attention,\\Flagged Heads\end{tabular} & Scorer \\
\midrule
32K & 10.9 & 12.1 & 12.2 & 0.74 & 0.94 & 0.98 & 1.06 \\
64K & 14.4 & 14.0 & 14.4 & 2.98 & 2.25 & 2.40 & 2.67 \\
128K & 20.9 & 19.3 & 19.9 & 13.04 & 5.61 & 6.23 & 7.13 \\
\bottomrule
 
\end{tabular}
\end{table}

\end{document}